\documentclass[12pt]{article}
\usepackage{amsmath}
\usepackage{graphicx}
\usepackage{enumerate}
\usepackage{natbib}
\usepackage{url} 
\usepackage{mathrsfs}
\usepackage{amssymb}
\usepackage{graphicx}
\usepackage{bm}
\usepackage{epstopdf}
\usepackage{amsthm}
\usepackage{color}
\usepackage{caption,subcaption}
\usepackage{lineno}
\usepackage{multirow}
\usepackage{hyperref}
\usepackage{tabularx}
\usepackage{threeparttable}
\usepackage{threeparttablex}

\newcommand{\blind}{0}

\newcommand{\bomega}{{\bm \omega}}

\newcommand{\bX}{{\bm X}}

\newcommand{\bh}{{\bm h}}
\newcommand{\bx}{{\bm x}}

\newcommand{\bd}{{\bm d}}

\newcommand{\bv}{{\bm v}}

\newcommand{\bW}{{\bm W}}
\newcommand{\bw}{{\bm w}}

\newcommand{\bbeta}{\boldsymbol{\beta}}
\newcommand{\btheta}{\boldsymbol{\theta}}

\newcommand{\balpha}{\boldsymbol{\alpha}}

\newcommand{\bgamma}{\boldsymbol{\gamma}}

\newcommand{\bSigma}{\boldsymbol{\Sigma}}

\newcommand{\bxi}{\boldsymbol{\xi}}
\newcommand{\biota}{\boldsymbol{\iota}}

\begin{document}

\def\spacingset#1{\renewcommand{\baselinestretch}%
{#1}\small\normalsize} \spacingset{1}


\if0\blind
{
\date{}
  \title{\bf Interpretable  Deep Regression Models with Interval-Censored Failure Time Data}
  \author{Changhui Yuan\\
    School of Mathematics, Jilin University\\
    and \\
    Shishun Zhao \\
    School of Mathematics, Jilin University\\
    and \\
    Shuwei Li  \\
    School of Economics and Statistics, Guangzhou University\\
    and \\
    Xinyuan Song \\
    Department of
    Statistics, Chinese University of Hong Kong\\
    and \\
    Zhao Chen  \\
    School of Data Science, Fudan University}
  \maketitle
} \fi

\bigskip
\begin{abstract}
Deep neural networks (DNNs) have become powerful tools for modeling complex data structures through sequentially integrating simple functions in each hidden layer. In survival analysis, recent advances of DNNs primarily focus on enhancing model capabilities, especially in exploring nonlinear covariate effects under right censoring. However, deep learning methods for interval-censored data, where the unobservable failure time is only known to lie in an interval, remain underexplored and limited to specific data type or model. This work proposes a general regression framework for interval-censored data with a broad class of partially linear transformation models, where key covariate effects are modeled parametrically  while nonlinear effects of nuisance multi-modal covariates are approximated via DNNs, balancing interpretability and flexibility. We employ sieve maximum likelihood estimation by leveraging monotone splines to approximate the cumulative baseline hazard function. To ensure reliable and tractable estimation, we develop an EM algorithm incorporating stochastic gradient descent. We establish the asymptotic properties of parameter estimators and show that the DNN estimator achieves minimax-optimal convergence. Extensive simulations demonstrate superior estimation and prediction accuracy over state-of-the-art methods. Applying our method to the Alzheimer's Disease Neuroimaging Initiative dataset yields novel insights and improved predictive performance compared to traditional approaches.
\end{abstract}

\noindent%
\vfill

\newpage
\spacingset{1.9} 
\section{Introduction}
\label{sec:intro}

In recent decades,  neural networks have exhibited remarkable success in various fields, including  medical image analysis \citep{sun2024penalized},  natural language processing \citep{devlin2018bert}, and object detection \citep{ren2016faster}.
A typical neural network  consists of  an input layer, one or more hidden layers, and an output layer.
By leveraging  an activation function, operation at each neuron is a nonlinear transformation for the simple weighted sum of the previous layer's outputs.
Neural networks have been theoretically recognized as
a remarkable   tool to  approximate unknown functions.
For example, \cite{cybenko1989approximation} and \cite{hornik1989multilayer} showed that
shallow neural networks can approximate any continuous function with any degree of accuracy, even with a single hidden layer.
A deep neural network (DNN) with multiple hidden layers and few nodes per layer
can attain better representational capability than a shallow neural network  with fewer than exponentially many nodes, which is an appealing advantage of DNN  \citep{telgarsky2015representation}.


In survival analysis,
neural networks have  spawned vast variants of the classical Kaplan-Meier estimator and regression models.  
To name a few,
\cite{katzman2018deepsurv} modeled  the complex interactions between  covariates and therapies with  a DNN in personalized treatment recommendations.
\cite{ren2019deep} investigated more flexible DNN-based Kaplan-Meier estimators.
\cite{kvamme2019time} developed   time-to-event prediction methods  by adopting the neural network versions  of the  proportional hazards (PH) model.
\cite{tong2022deep} proposed a nuclear-norm-based imputation method, extending the work of \cite{katzman2018deepsurv}  to handle missing covariates.



Recently, motivated by  DNN's powerful approximation ability,  nonparametric and partially linear regression models have revived in survival analysis  \citep{Xie2021SIM,   zhong2022deep, sun2023neural, wu2024deep}.
In particular,  partially linear models attract much attention since they can flexibly model linear and nonlinear covariate effects on the failure event of interest  \citep{FanCox1997, Huang1999,  ChenJin2006, lu2010estimation,  lu2018partially}.
Traditional estimation methods generally focus on constructing kernel-based estimating equations or using sieves, such as splines and local polynomials, to approximate the nonlinear component of the model. When the dimension of nonlinear covariates is large, the methods above suffer from the curse of dimensionality, which hinders their practical utility.
In contrast,  as shown by \cite{schmidt2020nonparametric} and \cite{bauer2019deep},
DNN can remedy the curse of dimensionality that often appears in nonparametric regressions, and the resulting estimators can achieve the optimal minimax convergence rate.
For  traditional  right-censored data,   \cite{zhong2022deep} provided a  partially linear PH model with DNN representing the nonlinear function of covariates, circumventing the curse of dimensionality and rendering a clear interpretation of interested covariate effects on a survival endpoint.


Interval-censored failure time data  mean that the failure time of interest (e.g.,  the  onset time of an asymptomatic disease) cannot be observed precisely and is only known to lie in  a time interval formed by periodic examinations or cross-sectional screening.
This type of censored  data is ubiquitous in   scientific studies, including but not limited to clinical trials, epidemiological surveys and sociological
investigations. A simple type of interval-censored data investigated extensively is called current status data, where each failure time is either larger or smaller than one observation time \citep{sun2006statistical}.
The analysis of interval-censored data is essential, and many studies have emerged  \citep{huang1996efficient,zhang2010spline, lu2018partially,LeeCY2022,LiPeng2023,Yuan2024SIM}.
For example,  \cite{wang2016flexible} and \cite{zeng2016maximum} proposed maximum likelihood estimation procedures for semiparametric PH and transformation models.
Despite a rich literature on interval-censored data,
exploring  DNN's utility  in partially linear models  is relatively limited.
The only  available methods were  \cite{wu2024deep} and \cite{du2024deep}, 
which  concerned sieve maximum likelihood analysis  of current status  and interval-censored data, respectively.
It is worth pointing out that  the two aforementioned methods were built upon the simple PH model,  in which the PH assumption may be fragile, especially in long-term studies involving chronic diseases
\citep{zeng2006}.

Exploring DNN-based models and associated reliable estimation procedures is essential to offering a highly flexible regression framework for analyzing interval-censored data. Besides enhancing the model capability and prediction accuracy, such an exploration can render reliable inferences by positing weak model assumptions.
The present study investigates this critical problem through a general  DNN-based partially linear transformation model. It encompasses the popular partially linear PH and proportional odds (PO) models as specific cases and is highly explanatory while maintaining sufficient flexibility. We employ monotone splines and DNN to approximate the cumulative baseline hazard function and other nonparametric components. We resort to sieve maximum likelihood estimation and develop an expectation-maximization (EM) algorithm with stochastic gradient descent (SGD) optimization embedded in the maximization step. The proposed algorithm is robust to initialization, which substantially relieves the burden of parameter tuning in training the DNN-based model, and yields the sieve estimators in a reliable and tractable way. Under certain smoothness and structure assumptions commonly used in the analysis of nonparametric models, the asymptotic properties of our proposed estimator are established. In particular,  the DNN-based estimator is shown to achieve the minimax optimal convergence. These properties ensure that the model is well-behaved and the estimator is consistent and efficient, providing a solid theoretical foundation for our approach.


\section{Methods}
\label{sec:model}
\subsection{Model,  Data Structure and Likelihood}

Let $\bX$ be a $p$-dimensional  covariate vector whose effect is of   primary interest (e.g., treatment variables).
Denote by $\bW$ a vector consisting of  $d$ nuisance covariates.
Conditional on $\bX$ and $\bW$, we posit that  the failure time of interest
$T$  follows a  partially linear transformation model, where  the  conditional cumulative hazard function of $T$ takes the form
\begin{equation}
	\Lambda (t \mid \bX, \bW) = G \left[ \Lambda (t) \exp\{  \bbeta ^{\top}  \bX + \phi(\bW)\}\right].
	\label{model}
\end{equation}
In model \eqref{model},
$\Lambda(t)$ represents an unspecified cumulative baseline  hazard function, $\bbeta$ is a $p$-dimensional vector of regression parameters,
$\phi (\bW)$ is an  unknown smooth function that maps $\mathbb{R}^d$ to $\mathbb{R}$,
and	$G(\cdot)$ is a prespecified increasing transformation function with
$G(0) = 0$.
Model  \eqref{model} is  general and includes various popular partially linear models as specific instances.
For example,
model  \eqref{model} corresponds to the partially linear PH and PO models
when setting $G(x)=x$ and $G(x) = \log(1+x)$, respectively.
In practical applications,  it is common to consider the logarithmic transformation family,
$G(x) = {\rm log} (1+r\,x) / r (r \geq 0)$, in  which  $r=0$ means $G(x)=x$.
In fact, this class of transformation functions can be derived from the Laplace transformation as follows
$G(x) = -\log \int_{0}^{\infty} \exp(-x \eta) f(\eta \mid r) {\rm d} \eta,$
where $f(\eta \mid r)$
is the density function of a gamma-distributed frailty variable with mean $1$ and variance $r$  \citep{kosorok2004robust}.
Since any constant in   $\phi (\cdot)$  can be absorbed into $\Lambda(\cdot)$, we assume that $E \{\phi ( \bW)\}=0$, a frequently used condition as in  \cite{zhong2022deep} and others, to ensure an identifiable model.


Notably,  traditional partially linear models,
such as the partially linear additive PH and transformation models in \cite{Huang1999} and  \cite{Yuan2024SIM},
typically assume an additive form,  $\sum_{j=1}^d \phi_j(W_j)$, for $\phi(\bW)$,
where  $W_j$ is the $j$th component of  $\bW$,
$\phi_j$ is an unknown function with $j = 1, \ldots, d$.
Unlike the additive modeling described above,  the proposed model \eqref{model}
leaves the form of  $\phi(\bW)$  completely unspecified
and can accommodate multi-modal $\bW$, thereby offering high flexibility.
Furthermore, it is easy to see that
model \eqref{model} can be equivalently expressed as
$$
\log \Lambda(t) =  -\bbeta^{\top}  \bX - \phi(\bW) + \epsilon,
$$
where  the error term $\epsilon$ has a distribution function $1 - \exp[-G\{\exp(x)\}]$.
Therefore,  $\bbeta$ essentially represents the effect  of  $\bX$ on the transformed $T$.

We consider general interval-censored failure time data,
where  the  exact failure time of interest $T$ cannot be obtained but is only known to lie in a specific  interval.
More specifically, let  $(L, R]$ be the interval that brackets $T$ with $L < R$.
Clearly, $T$ is left censored if $L = 0$ and right censored when $R = \infty$.
Define   $ \delta_{L} = 1$ if  $T$ is left censored,   $ \delta_{R} =1 $ when  $T$ is right censored, and  $ \delta_{I} =1$ when $T$ is strictly interval-censored (i.e.,  $L>0$ and  $R  < \infty$).
Then, the observed interval-censored data consist of $n$ i.i.d. realizations of
$\{L, R, \delta_{L}, \delta_{I}, \delta_{R}, \bX, \bW\}$,
denoted by  $\bm{O} = \{L_{ i }, R_{ i }, \delta_{ L , i }, \delta_{ I , i }, \delta_{ R , i }, \bX_{ i }, \bW_{ i };  i = 1,  \ldots , n\}$.
Notably,  $ \delta_{ L , i } + \delta_{ I , i } + \delta_{ R , i } =1 $ for each individual $i$.

Under the conditional independence assumption between  $T_i$ and   $(L_i, R_i)$ given $\bX_{ i }$ and $\bW_{ i }$,  the observed data likelihood   is given by
\begin{equation}\label{obsLik}
	\begin{aligned}
		\mathcal{L} _ { obs }& (\bbeta, \Lambda,\phi)\\
		&= \prod _ { i = 1 } ^ { n } F ( R _ { i } \mid \bX_i, \bW_i ) ^ { \delta _ { L,i } } \{ F ( R _ { i } \mid \bX_i, \bW_i) - F ( L _ { i } \mid  \bX_i, \bW_i ) \} ^ { \delta _ {I,i } }  \{ 1 - F ( L _ { i } \mid \bX_i, \bW_i) \} ^ { \delta _ {R,i } },
	\end{aligned}
\end{equation}
where $F ( t  \mid  \bX_i, \bW_i) = 1 - \exp  \left( - G \left[  \Lambda (t) \exp\{  \bbeta ^{\top}  \bX_i + \phi(\bW_i)\}\right]  \right)$
is the distribution function of $T_{ i }$   given $\bX_{ i }$ and $\bW_{ i }$.

Note that	the likelihood \eqref{obsLik} involves two infinite-dimensional
functions $\Lambda (t)$ and  $\phi ( \bW)$,
which are usually treated as nuisance functions in model \eqref{model}.
Since  $\Lambda (t)$ is increasing and satisfies $\Lambda (0) = 0$,
it is routine to adopt a sieve approach, which approximates $\Lambda (t)$ by  some surrogate function  with a finite number of  parameters.
Herein,  we consider employing monotone splines \citep{Ramsay1988}, rendering the following  approximation:
$$\Lambda ( \cdot ) \approx \Lambda _ {\bgamma} ( \cdot ) = \sum _ { l = 1 } ^ {L_n} \gamma _ { l } M _ { l } ( \cdot ), $$
where  $ \gamma _ { l }$'s are the  nonnegative spline coefficients that ensure the  monotonicity of  $\Lambda _ { \gamma } ( \cdot )$, $ M_{l}( \cdot )$'s  represent the integrated spline  basis functions, each of
which is nondecreasing from 0 to 1, and
$ \bgamma = (\gamma_{1} , \ldots,  \gamma_{L_n})^{\top} $.
In order to use   monotone splines, one needs to  specify the degree  denoted by $d_M$ and an interior knot set specified within a time range.
In particular, the overall smoothness of the
basis functions is determined by the degree.
For instance,  monotone splines with  degrees of 1, 2, or 3 correspond  to linear, quadratic, or cubic splines, respectively.
More dense interior knots in a time range  offer more flexible  modeling in that range.
In practical applications, the interior knots can be set to be an equally spaced grid of points between the minimum and maximum observation times or quantiles of the observation times.
After specifying the degree  and   interior knots,
one can determine
$L_n = p_n + d_M$ spline basis functions, where  $p_n$ is the interior knot  number.

\subsection{Deep Neural Network}

To estimate   $\phi (\bW)$,
traditional methods,  such as kernel smoothing,
generally suffer from the curse of dimensionality especially when encountering a large-dimensional $\bW$.
To circumvent this obstacle,  we herein consider
approximating 	$\phi (\bW)$
with the powerful DNN  denoted by $\phi_{\balpha} (\bW)$, where $\balpha$ is a vector consisting of all unknown parameters in  the DNN.
Let $q$ be a positive integer representing the number of hidden layers in the DNN.
A typical $(q + 1)$-layer DNN  consists of an input layer,  $q$ hidden layers, and an output layer.
Denote  by $\bh  = (h_0, h_1, \ldots, h_q, h_{q+1})$   some positive integer sequence, where $h_0$ is the dimension of input variables in the input layer, $h_j$ is the number of neurons for  the $j$th hidden layer with $j = 1, \ldots, q$, and $h_{q+1}$ is  the dimension of the output layer.
A  $(q + 1)$-layer
DNN with layer-width $\bh$   is essentially a composite function $\phi_{\balpha}$:  $\mathbb{R}^{h_{0}} \rightarrow \mathbb{R}^{h_{q+1}}$
recursively defined as
\begin{equation}
	\begin{aligned}
		\phi_{\balpha}(\bx)  &= \bomega_{q}\phi_{q}(\bx)  + \bv_{q}, \\
		\phi_q(\bx)  &= \sigma(\bomega_{q-1}\phi_{q-1}(\bx)  + \bv_{q-1}), \\
		\vdots &\\
		\phi_2(\bx)  &=  \sigma(\bomega_1 \phi_1(\bx) + \bv_1) ,\\
		\phi_1(\bx)  &=  \sigma(\bomega_0  \bx + \bv_0),
		\label{phi_NN}
	\end{aligned}
\end{equation}
where $\bomega_0, \ldots, \bomega_{q}$ are the unknown weight matrices, and $\bv_0, \ldots, \bv_{q}$ are the unknown vectors
including the shift term (i.e., intercept).
For each $j=0, 1, \ldots, q$, $ (\bomega_j)_{s,k}$,  the $(s,k)$th entry of $\bomega_j$,
denotes the weight associated with the $k$th neuron in layer $j$ to the $s$th neuron in
layer $j + 1$, and the vector entry $(\bv_j)_s$ represents a shift term associated with the $s$th neuron
in layer $j + 1$.
The activation functions denoted by $\sigma$
perform  component-wise nonlinear transformations on vectors as
$\sigma((x_1, \ldots, x_{h_j})^{\top}) = (\sigma(x_1), \ldots, \sigma(x_{h_j}))^{\top}$ for $j = 1, \ldots, q$.
Thus,   for each $j = 1, \ldots, q$,   $\phi_j = (\phi_{j1}, \ldots, \phi_{j{h_j}})^{\top}$ maps $\mathbb{R}^{h_{j-1}}$ to $\mathbb{R}^{h_{j}}$.
For illustration,
Figure 1 presents the network structure of a 4-layer DNN with $\bh = (h_0, h_1, h_2, h_3, h_{4}) = (4, 3, 3, 2, 1)$.
\begin{figure}
	\begin{center}
     \includegraphics[ width=6in]{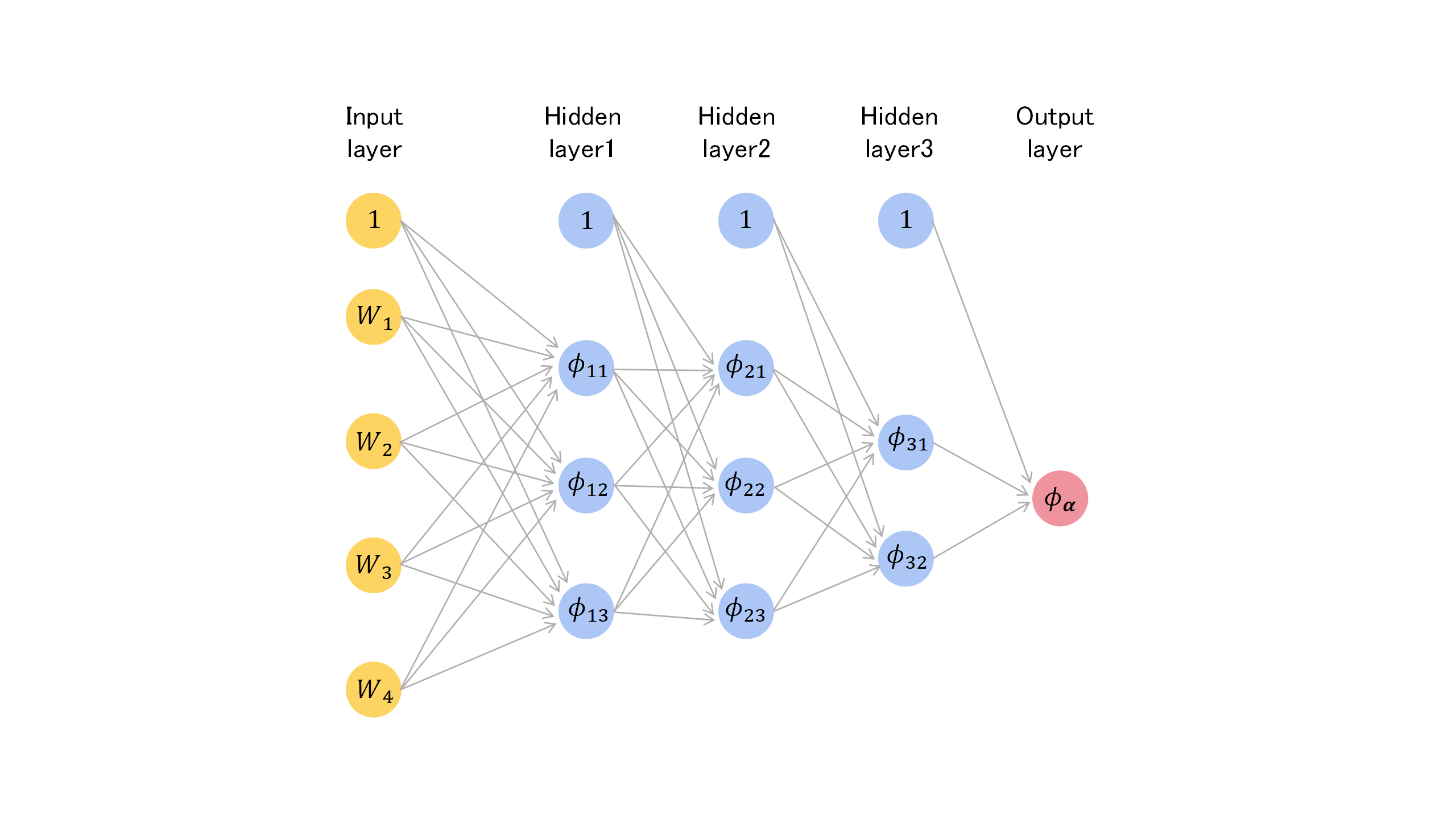}
	\end{center}
	\caption{The network structure of a 4-layer DNN with $\bh = (h_0, h_1, h_2, h_3, h_{4}) = (4, 3, 3, 2, 1)$.The filled circle over the neurons of the input layer or each hidden layer implies the existence of a  shift term. \label{fig:first}}
\end{figure}

Some commonly adopted activation functions in deep learning include sigmoid, tanh, rectified linear unit (ReLU), and their variants \citep{he2015delving,chen2020arelu}.
In our DNN architecture, the Scaled Exponential Linear Unit (SeLU) \citep{klambauer2017self} is the activation function. The SeLU activation function can be considered a variant of the simple ReLU and has several advantages over ReLU.
For example,
unlike ReLU, which outputs zero for negative inputs, SeLU has a non-zero output in the negative region,  keeping potentially important neurons active in the whole network architecture \citep{dubey2022activation}.
Therefore, adopting the SeLU activation function can lead to stable and efficient training of DNN.

Note that, in a
DNN, if neurons in any  two adjacent layers  are fully connected, the dimension of $\balpha$ would be large, which may lead to overfitting.
In practical implementations,  pruning weights such that the network's layers are only sparsely connected
is a routine to relieve overfitting \citep{silver2017mastering,schmidt2020nonparametric}.
Specifically, define a class of sparse neural networks as
$$
\begin{aligned}
	\mathcal{F}(s,q,\boldsymbol{h},B) =\Big\{\phi_{\balpha}: \,\, & \phi_{\balpha} \,\, \text{is a DNN with $(q + 1)$ layers and width vector $\boldsymbol{h}$ such that}\\
	& \max\{ \|\bomega_j\|_{\infty},  \|\bv_j\|_{\infty}\} \leq 1 \,\, \text{for} \,\,  j = 0, \ldots, q,     \\
	&\sum_{j=0}^q (\|\bomega_j\|_0 + \|\bv_j\|_0) \leq s, \|\phi_{\balpha}\|_\infty\leq B \Big\},
\end{aligned}
$$
where $s \in \mathbb{R}_+$,  the set of all positive real numbers,  $B>0$,
$\|\cdot\|_{\infty}$ denotes the sup-norm of a matrix, vector, or function, and $ \|\cdot\|_0$  is the number of nonzero entries of a matrix or vector.
The sup-norm of a function $\phi$ is defined as $\|\phi\|_\infty = \sup_{\bW \in \mathbb{R}^d} |\phi (\bW)|$.
Typically,  the sizes of the learned matrices $\bomega_j$'s and vectors $\bv_j$'s are not large, especially when the initial matrices and vectors used for starting the SGD training are relatively small.
Thus, as in \cite{schmidt2020nonparametric} and others, we assume that  the unknown parameters in the DNN  are  bounded by one.	
In a DNN, the parameter $s$ governs the network sparsity, and its selection typically leverages the dropout technique within each hidden layer, which randomly ignores partial neurons based on a specified dropout rate  \citep{srivastava2014dropout}.
This strategy enhances model robustness by preventing over-reliance on particular neurons during training \citep{sun2024penalized}.
In model \eqref{model},  by approximating $\phi(\bW)$ with a  DNN $\phi_{\balpha} \in  \mathcal{F} = \mathcal{F}(s,q,\boldsymbol{h}, \infty)$, where
$h_0 = d$ and $h_{q+1}=1$,
we have the proposed deep partially linear transformation model.

\subsection{Parameter Estimation}

Based on the aforementioned spline and DNN approximations, the likelihood  \eqref{obsLik} can be rewritten as
\begin{equation}
	\begin{aligned}
		\mathcal{L}& _ { obs } (\bbeta, \bgamma, \balpha) \\
		&= \prod _ { i = 1 } ^ { n } F_n ( R _ { i } \mid \bX_i, \bW_i ) ^ { \delta _ { L,i } } \{ F_n ( R _ { i } \mid \bX_i, \bW_i) - F_n ( L _ { i } \mid  \bX_i, \bW_i ) \} ^ { \delta _ {I,i } } \{ 1 - F_n ( L _ { i } \mid \bX_i, \bW_i) \} ^ { \delta _ {R,i } },
		\label{obsLik-22}
	\end{aligned}
\end{equation}
where $F_n ( t  \mid  \bX_i, \bW_i) = 1 - \exp  \left( - G \left[ \Lambda _ {\bgamma} ( t ) \exp\{  \bbeta ^{\top}  \bX_i + \phi_{\balpha} ( \bW_i) \} \right]  \right)$.

By the maximum likelihood principle,   	
the estimator of $(\bbeta, \bgamma, \balpha)$ denoted by $(\hat{\bbeta}, \hat{\bgamma}, \hat{\balpha})$ is defined as
$$
(\hat{\bbeta}, \hat{\bgamma}, \hat{\balpha}) =  \mathop{\arg \max}_{\bbeta \in \mathbb{R}^p, \bgamma \in \mathbb{R}_+^{L_n},   \phi_{\balpha} \in  \mathcal{F}} \,\,\,  \mathcal{L} _ { obs } (\bbeta, \bgamma, \balpha).
$$
Typically, from the computational aspect,
performing direct maximization of likelihood \eqref{obsLik-22} with some existing optimization functions
in the statistical software is quite challenging
due to the intractable form of \eqref{obsLik-22} and many unknown parameters.
The complexity is further magnified by including  a neural network estimation.
Such an optimization is not a simple task, even without the DNN $\phi_{\balpha}(\cdot)$  under a simpler PH model \citep{wang2016flexible}.

To facilitate the estimation, we construct an EM algorithm coupled with a three-stage data augmentation and SGD  to maximize \eqref{obsLik-22} and obtain $(\hat{\bbeta}, \hat{\bgamma}, \hat{\balpha})$.
Due to the space restriction, we defer the details of the proposed algorithm to Section S.1 of the supplementary materials.
Empirically, the proposed algorithm has  certain  robustness  to the choice of initial values,
$\bbeta^{(0)}$, $\bgamma^{(0)}$, and $\balpha^{(0)}$.
In practice, one can initialize the monotone spline coefficients with 0.01 and all regression coefficients with 0.
We employ Keras's default random initialization for  all parameters in the DNN \citep{glorot2010understanding}.
Convergence of the algorithm is declared when the maximum absolute difference of the log-likelihood values between two successive iterations is less than a small positive threshold (e.g., 0.001).

\section{Asymptotic Properties}
\label{sec:AsyP}
We establish the asymptotic properties of our proposed estimators with
the sieve estimation theory and empirical process techniques.
Let $\hat{\btheta}= (\hat{\bbeta}, \hat{\Lambda}, \hat{\phi})$ denote the proposed estimator of $\btheta= (\bbeta, \Lambda, \phi)$,
where $\hat{\bbeta}$,  $\hat{\Lambda}$, and $\hat{\phi}$ are the estimators of $\bbeta$,  $\Lambda$, and $\phi$, respectively.
Define $\btheta_{0} = (\bbeta_{0}, \Lambda_{0}, \phi_{0})$,
where $\bbeta_0$,  $\Lambda_0$, and $\phi_0$ represent the true values of $\bbeta$,  $\Lambda$, and $\phi$,  respectively.
Let $ \| \cdot \|$ and $ \| \cdot \|_{\infty} $ denote the usual Euclidean  norm and supremum norm, respectively.
That is, for a vector $\bbeta = (\beta_1, \ldots, \beta_p )^{\top},  \|\bbeta\|=(\sum_{i=1}^{p}\beta_i^2)^{1/2}$ and  $\|\bbeta\|_{\infty}= \max_i  | \beta_i |$.
For any matrix $\bSigma = (\sigma_{ij}), \|\bSigma\|_{\infty}= \max_{i,j}  | \sigma_{ij} |$.
The notation  $X_n \lesssim Y_n$ indicates  $X_n \leq MY_n$, $X_n \gtrsim Y_n$ indicates  $X_n \geq MY_n$ for some constant $M>0$.
$X_n \asymp Y_n$  indicates $X_n \lesssim Y_n$  and $X_n \gtrsim Y_n$, which means that $X_n$ is of the same order as $Y_n$.
Define  $X_n \wedge Y_n = \min\{X_n, Y_n \}$ and  $X_n \vee Y_n = \max\{X_n, Y_n \}$.
Let   $\mathbb{N}_+$  denote  the   set of all positive integers   and
$\mathbb{R}_+$ represent  the set of all positive real numbers.
Define a H\"{o}lder class of smooth functions with  $\xi_{\phi} >0$, $S>0$  and $\mathbb{D}\subset\mathbb{R}^d$ as
$$\Psi_d^{\xi_{\phi} }(\mathbb{D},S)=\left\{\phi:\mathbb{D}\to\mathbb{R}:\sum_{\biota:|\biota|<\xi_{\phi} }\left\|\partial^{\biota} \phi\right\|_\infty +\sum_{\biota:|\biota|=\lfloor \xi_{\phi}  \rfloor}\sup_{\bw_1, \bw_2\in\mathbb{D}, \bw_1\neq \bw_2}\frac{|\partial^{\biota} \phi(\bw_1)-\partial^{\biota} \phi(\bw_2)|}{\|\bw_1-\bw_2\|_\infty^{\xi_{\phi} -\lfloor \xi_{\phi}  \rfloor}}\leq S\right\},$$
where we use the multi-index notation	$\partial^{\biota} = \partial^{\iota_1} \ldots\partial^{\iota_{d }}$ with $\biota= (\iota_1,\ldots,\iota_d)^{\top} \in \mathbb{N}_+^d$  and $|\biota|= \sum_{j=1}^d \iota_j$,
and $\lfloor \xi_{\phi}  \rfloor$ is the largest integer that is strictly smaller than $\xi_{\phi}$  \citep{schmidt2020nonparametric}.

Let $\tilde{k} \in\mathbb{N}_+$ and  $\bxi_{\phi}=(\xi_{\phi1},\ldots,\xi_{\phi {\tilde{k}}})^{\top} \in\mathbb{R}_+^{\tilde{k}}$.
Define $\bar{\bd}=(\bar{d}_1,\ldots,\bar{d}_{\tilde{k}+1})^{\top} \in\mathbb{N}_+^{\tilde{k}+1}$
and $\tilde{\bd}=(\tilde{d}_1,\ldots,\tilde{d}_{\tilde{k}})^{\top} \in\mathbb{N}_+^{\tilde{k}}$ with $\tilde{d}_j \leq \bar{d}_j $ for each $j=1,\ldots,\tilde{k}$.
Let $\tilde{\xi }_{\phi i} = \xi_{\phi i} \prod_{j=i+1}^{\tilde{k}}(\xi_{\phi j} \wedge 1)$ for $i = 1, \ldots, \tilde{k}$, and $\alpha_n = \max_{i= 1, \ldots \tilde{k}} n^{-\tilde{\xi }_{\phi i} /(2\tilde{\xi }_{\phi i} +\tilde{d_i } )}$.
We 	further  define a composite smoothness function class, taking the form
\begin{flalign}
	\Psi(\tilde{k},\bxi_{\phi}, \bar{\bd},\tilde{\bd},S) =\Bigg\{ & \phi=\varphi_{\tilde{k}}\circ\cdots\circ \varphi_1:\varphi_i=(\varphi_{i1},\ldots,\varphi_{i \bar{d}_{i+1}})^{\top} \text{ and}\Bigg.\\  \nonumber
	&\Bigg.
	\varphi_{ij} \in \Psi_{\tilde{d}_i}^{\xi_{\phi i}}([e_i,f_i]^{\tilde{d}_i},S),   \text{for some} \,  |e_i|  \,  \text{and} \,  |f_i| \leq  S,  i = 1, \ldots, \tilde{k}, \Bigg.\\  \nonumber
	&\Bigg.  j = 1, \ldots, \bar{d}_{i+1} \Bigg\},
\end{flalign}
where    $S>0$ and $\tilde{\bd}$ represents the intrinsic dimension of a function.
For instance, if  $\bw\in [ 0, 1] ^{10}$,
$$
\phi(\bw) = \varphi_{31}\big( \varphi_{21}\{\varphi_{11}( w_{1}, w_{2}, w_3, w_{4}) , \varphi_{12}(w_{5},w_6) \}, \varphi_{22}\{\varphi_{13}( w_{7}) ,\varphi_{14}( w_{8}) , \varphi_{15}( w_{9}, w_{10})\}\big),
$$
and
each $\varphi_{ij}$ is  twice continuously differentiable, then  $\bxi_{\phi}=(2,2,2)$, $\tilde{k}=3$, $\bar{\bd}= (10,5,2,1)$  and $\tilde{\bd}=(4,3,2).$
Let
$\ell^{\infty}[x,y]$ denote the space of bounded sequences on $[x,y]$.
Define the function   spaces  $\bar{\Psi}_{\Lambda} = \{\Lambda \in \ell^{\infty}[a,b]: \Lambda $ is monotone increasing with $\Lambda(0)=0$ and $\Lambda(b) < \infty$,
where $[a,b]$ is the union   support  of $L$ and $R$ with $ 0 \leq a < b < \infty\}$
and   $\bar{\Psi}_{\phi} = \{ \phi:  \phi \in 	\Psi(\tilde{k},\bxi_{\phi},\bd,\tilde{\bd},S),  E \{\phi ( \bW)\}=0 \}$.

Let $\boldsymbol{\Theta}$ be the product space defined as $\boldsymbol{\Theta}=  \mathcal{D} \times \bar{\Psi}_{\Lambda} \times \bar{\Psi}_{\phi}$, where $\mathcal{D}$ is a compact subset of $\mathbb{R}^p$.
For any $\btheta_1$ and  $\btheta_2 \in \boldsymbol{\Theta} $,  define the distance $d(\btheta_1,\btheta_2)$ as
$$
d(\btheta_1,\btheta_2) =  ( \| \bbeta_{1} - \bbeta_{2} \|^{2} + \| \Lambda_{1} -  \Lambda_{2} \|_{L_{2}}^{2} + \| \phi_{1} - \phi_{2} \|_{L_{2}}^{2} )^{1/2},
$$
where $\| \cdot \|_{L_{2}}$  denotes the $L_2$-norm
over a specific interval.
For instance,  $\left\|\phi_1-\phi_2\right\|^2_{L_{2}}=\int_{\mathcal{W}}\{\phi_1(\bw)-\phi_2(\bw)\}^2 \mathrm{~d} F_{\bW}(\bw)$,  where $F_{\bW}(\bw)$ is the distribution function of $\bW$,
$\mathcal{W}$ is the support of $\bW$, $\phi_1 \in \bar{\Psi}_{\phi}$ and   $\phi_2 \in \bar{\Psi}_{\phi}$.

Consider the set  $\mathcal{T}_n=\{t_i\}_{i=1}^{p_n}$, where
$$
a=t_{0}<t_{1}<\ldots<t_{p_n}<t_{p_n+1}=b
$$
forms a sequence of knots dividing the interval $[a, b]$  into $p_n+1$ subintervals, and $p_n=O(n^\nu)$   with  $0<\nu<0.5$.
Define
$$
\mathcal{M}(\mathcal{T}_{n}, d_M, B) = \left\{ \sum_{l=1}^{L_n} \gamma_{ l }M_l(t) : 0 \leq \gamma_{ l } \leq  B   \text{ for }  l=1, \ldots, L_n, t \in [a,b] \right\},
$$
where $B$ is a large positive constant.
To establish  the asymptotic properties of $\hat{\btheta}$,  we use the following regularity conditions.
\begin{enumerate}[{\bf(C1)}]
	\item
	The true regression parameter vector $ \bbeta_{0} $ belongs to the interior of the compact set $\mathcal{D} $,  $\Lambda_{0} \in \bar{\Psi}_{\Lambda}$,  the first derivative of $\Lambda_{0}$ is strictly positive and continuous on $[a, b]$,  and $\phi_0 \in  	\bar{\Psi}_{\phi}$.
	\item
	The covariate vector $(\bX^{\top}, \bW^{\top})^{\top}$ takes values in a bounded subset of $\mathbb{R}^{p+d}$, and its  joint probability density function is bounded away from zero.
	\item
	For $ \xi  \geq 1 $,  the $\xi $th derivative  of $ \Lambda_{0} $ satisfies the Lipchitz condition  on $ [a, b] $. That is,   there exists a constant $c_{\Lambda}$ such that
	$\lvert\Lambda_{0}^{(\xi)}(t_1) - \Lambda_{0}^{(\xi)}(t_2) \rvert\leq  c_{\Lambda}\lvert t_1-t_2 \rvert,$
	for any $t_1,t_2\in [a,b]. $
	\item
	The maximum spacing  between any two adjacent  knots,
	$\Delta_{\Lambda} = \max\limits_{1 \leq i \leq p_n+1}\lvert t_i - t_{i-1}\rvert$,
	is   $ O(n^{-\nu})$ with $0 < \nu < 0.5$.
	In addition, the ratio  of the maximum and minimum spacings  of the adjacent  knots,  $\Delta_{\Lambda}/\delta_{\Lambda}$,  is  uniformly  bounded, where $\delta_{\Lambda} = \min\limits_{1 \leq i \leq p_n+1}\lvert t_i - t_{i-1}\rvert$.
	\item For any $\bbeta \neq \bbeta_{0},  P(\bbeta^\top \bX \neq \bbeta_0^\top \bX) >0 $.
	\item In the DNN,  $q = O(\log n), s = O(n \alpha_n^2\log n)$, and $ n \alpha_n^2 \lesssim \min \{h_1, \ldots, h_q\} $\\
	$ \leq \max \{h_1, \ldots, h_q\} \lesssim n$.
	\item
	For $  \tilde{\xi}  \geq 1 $,  the joint density $ f(t, \bx, \bw, \delta_{l}, \delta_{i},  \delta_{r}) $ of $ (T,\bX, \bW, \delta_{L}, \delta_{I},  \delta_{R}) $ with respect to $  t $ or $ \bw $ has bounded $ \tilde{\xi}$th partial derivatives.
	\item
	If $\bbeta^{\top} \bX+f(t)=0$ for any $t \in [a, b]$  with probability one, then $\bbeta =0 $ and $f(t)=0$ for $t \in [a, b]$.
\end{enumerate}

Conditions (C1) and (C2)  impose boundedness  on the true parameters and covariates, which are commonly employed in the literature of interval-censored data analysis  \citep{huang1996efficient,zhang2010spline,zeng2016maximum}.
Condition (C3) pertains to the smoothness of $\Lambda_{0}(t)$, and
Condition (C4) is frequently adopted to  establish the convergence rate and asymptotic normality of a sieve estimator \citep{lu2007estimation}.
Condition (C5) pertains to  the model identification.
Condition (C6) specifies a neural network's structure (e.g., the sparsity and  layer-width) in $\mathcal{F}(s,q,\boldsymbol{h},B)$
and  serves as a balance between approximation error and estimation error  \citep{zhong2022deep}.
Condition (C7) is used to show that the score operators for nonparametric components trend toward zero in the least favorable direction, facilitating the asymptotic normality of parametric component estimators.
Condition (C8) holds if the matrix $E([1, X^{\top}]^{\top}[1, X^{\top}])$ is nonsingular.
The  asymptotic behaviors of $\hat{\btheta}$ are described in the  following three theorems
with the detailed  proofs  sketched in Section S.2 of
the supplementary materials.

\newtheorem{theorem}{\bf Theorem}

\begin{theorem}\label{thm1}
	Suppose that  conditions (C1)--(C5) hold,  we have
	$\| \hat{\bbeta} - \bbeta_{0} \|  \rightarrow 0$ and
	$ \| \hat{\Lambda} -  \Lambda_{0} \|_{L_{2}}  + \| \hat{\phi} - \phi_{0} \|_{L_{2}}  \rightarrow 0 $,  as  $n \rightarrow \infty$.
\end{theorem}

\begin{theorem}\label{thm2}
	Suppose that conditions (C1)--(C6) hold and  $ 1/(2\xi +1) < \nu < 1/(2\xi) $ with $\xi \geq 1 $, we have
	$d(\hat{\btheta}, \btheta_{0}) = O_{p}(\alpha_n \log^2n+n^{-\xi\nu}). $
\end{theorem}

\begin{theorem}\label{thm3}
	Suppose that conditions   (C1)--(C8) hold, then
	$$ n^{1/2}(\hat{\bbeta} - \bbeta_{0}) \stackrel{d}{\rightarrow} \mathcal{N}\{0, {\bf I}^{-1}(\bbeta_{0}) \}, \, {\rm as} \, n \rightarrow \infty,$$
	where  $\stackrel{d}{\rightarrow}$ denotes the convergence in distribution and   ${\bf I}(\bbeta_{0})$ is the information matrix of $\bbeta_{0}$ defined in  Section S.2 of
the supplementary materials. In other words, the covariance matrix of $n^{1/2}(\hat{\bbeta} - \bbeta_{0})$ attains the semiparametric efficiency
	bound.
\end{theorem}


To make inference on regression vector of interest $\bbeta_{0}$, one often needs to estimate the covariance matrix of $\hat{\bbeta}$.
Herein, we   employ a simple numerical profile likelihood approach.
Specifically,  define $ \mathcal{L}_i(\bbeta, \Lambda,\phi)$ as the logarithm of the observed-data likelihood function for subject $i$  in $\mathcal{L} _ { obs } (\bbeta, \Lambda,\phi) $ \eqref{obsLik}.
For a fixed $\boldsymbol{\beta}$, define $(\hat{\Lambda}_{\bbeta}, \hat{\phi}_{\bbeta})=\operatorname{argmax}_{\Lambda,\phi} \log \mathcal{L} _ { obs } (\bbeta, \Lambda,\phi)$,
which can be obtained via the proposed EM algorithm by   only updating  $\bgamma$ and $ \balpha$.
Thus, the covariance matrix of $\hat{\bbeta}$ is given by $(n \hat{\boldsymbol{I}})^{-1}$, where
$$
\hat{\boldsymbol{I}}=n^{-1} \sum_{i=1}^n\left\{ \left.\frac{\partial}{\partial \bbeta} \mathcal{L}_i\ (\bbeta, \hat{\Lambda}_{\bbeta}, \hat{\phi}_{\bbeta} )\right|_{\bbeta=\hat{\bbeta}}\right\}^{\otimes 2} .
$$
In the above, $\boldsymbol{a}^{\otimes 2}=\boldsymbol{a} \boldsymbol{a}^{\top}$ for a column vector $\boldsymbol{a}$.
For each $i$,  to calculate $ \frac{\partial}{\partial \bbeta} \mathcal{L}_i (\bbeta, \hat{\Lambda}_{\bbeta}, \hat{\phi}_{\bbeta} ) |_{\bbeta=\hat{\bbeta}}$,  one can readily employ a first-order numerical difference method given by
$$
\frac{p \ell_i\left(\hat{\boldsymbol{\beta}}+h_n e_j\right)-p \ell_i(\hat{\boldsymbol{\beta}})}{h_n},
$$
where $e_j$ is a $p$-dimensional vector with 1 at the $j$th position and 0 elsewhere, $h_n$ is a constant with the same order as $n_t^{-1 / 2}$, $ n_t $ represents the sample size of the training data, and $p \ell_i(\boldsymbol{\beta})=\mathcal{L}_i (\boldsymbol{\beta}, \hat{\Lambda}_{\bbeta}, \hat{\phi}_{\bbeta} )$.
Our numerical experiments suggest that this profile likelihood variance estimation method   works well in finite samples.


\section{Simulation Study}
\label{sec:simu}

We conducted extensive simulations to assess the empirical performance of the proposed deep regression method in finite samples. Comparison results with  the deep PH model   \citep{du2024deep}, the partially linear additive transformation model   \citep{Yuan2024SIM},  the penalized PH model with linear covariate effects  \citep{zhao2020simultaneous}, and the  transformation model  with linear covariate effects   \citep{zeng2016maximum} were also included to demonstrate the advantages of the proposed method.
In model  \eqref{model},  we let $\bX = (X_{1},  X_{2})^{\top}$, where $X_{1}$ and $X_{2}$ were generated from the standard normal distribution and the Bernoulli distribution  with a success probability of $0.5$, respectively.
The length of  $\bW$ was set to $d= 4$ or
$10$, and each component of  $\bW = (W_1, \ldots, W_d)^{\top}$ followed the uniform distribution on $(-1, 1)$.
The failure time $T$ was generated according to   model \eqref{model}, where  $\bbeta = (\beta_{1}, \beta_{2})^{\top}=(0.5, -0.5)^{\top}$, $G(x) =x$ or $\log(1 + x)$, $\Lambda (t)= 0.1t$ if $d=4$ and $\Lambda (t) =  0.2t$ if $d=10$.   	Notably, model \eqref{model} reduces to the deep partially linear  PH and PO models by specifying $G(x) =x$ and $G(x) = \log(1 + x)$, respectively.
For  the setting of  $d=4$, we investigated the following   three   cases for  $\phi(\bW)$:

Case 1: $\phi(\bW)= W_1 + W_2 /2 + W_3 /3 + W_4 /4,$

Case 2 : $\phi(\bW) = W_1^2+2W_2^2+W_3^3+\sqrt{W_4+1}-1.9,$	
and

Case 3 : $\phi(\bW) = W_1^2
+ \log (W_2+2) /2 + 2\sqrt{W_3W_4+1}-2.6$.

\noindent	
For the setting of  $d=10$,  the true $\phi(\bW)$ was specified as

Case 4: $\phi(\bW)= W_1 + W_2 /2 +\cdots + W_9  /9 + W_{10} /10,$

Case 5 : $\phi(\bW) = W_1+W_2+\cdots+W_8+W_9^2+2W_{10}^2-1,$ or

Case 6 : $\phi(\bW) = \log(W_1+1)+W_2^2W_3^3+W_4/2+\sqrt{W_5W_6+1}+W_7^2+\exp(W_8/2)+(W_9+W_{10})^2-2.7.$

\noindent	
Constants, -1.9, -2.6, -1.0,  and -2.7, were included in $\phi(\bW) $  to ensure that  $E\{\phi(\bW)\}$ is  close to 0.

To create interval censoring, we generated a sequence of observation times, where the gap between any two consecutive observations followed an exponential distribution with a mean of 0.5.
We set the study length to 8, after which no further observations happened.
Then, the interval $(L, R]$ that contains $T$ was determined by contrasting
the generated $T$ with the  observation times.
On average,  the above configurations gave approximately 5\% to 16\% left-censored observations and 34\% to 61\% right-censored ones.

To apply the proposed estimation method,  we approximated   $\Lambda(t)$ with cubic monotone splines, in which 3 interior knots were placed at the equally spaced quantiles within the minimum and maximum  observation times.  We set  the sample size $n$ to $500$ or $1000$ and used 200 replicates.
For each simulation replicate,
16\% of the simulated data were treated as the validation data used to tune hyperparameters,
including the layer number, learning rate, and tuning parameter of the $L_1$ penalty used in the SGD algorithm.
We then specified 64\% of the simulated data as the training set utilized to estimate the unknown parameters based on the selected optimal hyperparameter configuration.
Therefore,  the actual sample sizes used to obtain the parameter estimation results were 320 and 640 for $n = 500$ and 1000, respectively.
The remaining 20\% of the data served as test data to evaluate the estimation performance for $\phi(\bW)$.

We considered 2 or 3 hidden layers,
50 neurons in each hidden layer, and SeLU activation function in building the neural network.
To train the  network,  we set the tuning parameter in the $L_1$ penalty to  $0.01$ or  $0.05$ and  the learning rate to $0.0001$  or  $0.0003$.
The mini-batch size in the SGD algorithm was set to 50. The neural network was trained for 20 epochs during each iteration of  the proposed EM algorithm.
We applied the dropout regularization with a rate of 0.1 to prevent overfitting and selected the optimal hyperparameter combination with the maximum likelihood principle on the validation data.

Table \ref{table1}  summarizes the simulation results for the  regression parameter estimates under the PH model ($G(x)=x$).
Based on 200 replications, the used metrics include the estimation bias (Bias) calculated by the average of the estimates minus the true value,  the sample standard error (SSE) of the estimates,  the average of the standard error estimates (SEE), and the 95\% coverage probability (CP95) formed by the normal approximation. Results in Table \ref{table1} indicate that the proposed method gives sensible performance  in finite samples.
The biases of all regression parameter estimates are small, and the ratio of SSE to SEE is close to 1, suggesting that our profile-likelihood variance estimation approach performs adequately.
All coverage probabilities are close to the nominal value, affirming the validity of using a normal approximation for the asymptotic distribution of regression parameter estimates.
As anticipated, the proposed method's estimation efficiency exhibited by SSEs and SEEs improves when the sample size increases from 500 to 1000.

\begin{table}
	\caption{Simulation results for the regression parameter  estimates under the PH model. Results include	the estimation bias (Bias),  the sample standard error (SSE) of the estimates,  the average of the standard error estimates (SEE),  and   the 95\% coverage probability (CP95).}\label{table1}
	
	\begin{center}
		\scalebox{0.68}{
		\begin{threeparttable}
		\begin{tabular}{rrrrrrrrrrrrrrrrrrr}
			&        &       & \multicolumn{4}{c} {Proposed method}   & & \multicolumn{2}{c}{Deep PH}      & & \multicolumn{2}{c}{Spline-Trans} & & \multicolumn{2}{c}{ Penalized PH}& & \multicolumn{2}{c}{NPMLE-Trans }\\
			\cline{4-7} \cline{9-10}\cline{12-13}\cline{15-16}\cline{18-19}
			Case   & $n$ &  par. & Bias&SSE & SEE&CP95& & Bias&SSE & & Bias&SSE & & Bias&SSE& & Bias&SSE \\
			\hline
			1        & 500 & $\beta_{1}$& 0.011&0.086 &0.094 &0.985 &&  0.037    &  0.098  & &-0.002& 0.145& &-0.014& 0.086& &0.008& 0.086  \\
			&        &   $\beta_{2}$& -0.009&0.167  &0.176 &0.965&&   -0.020   &   0.186  & &-0.038& 0.186& & 0.069& 0.206 & &-0.008& 0.168   \\
			&1000 &  $\beta_{1}$& 0.018&0.059 &0.063 &0.965 &&  0.017 &  0.060  & &0.030& 0.061&  & 0.008& 0.059 & &0.023& 0.060 \\
			&   &  $\beta_{2}$& -0.015&0.125 &0.121 &0.965 &&    -0.010  &  0.125  & &-0.028& 0.134& &   0.011& 0.129 & &-0.021& 0.125  \\
			2        & 500 &$\beta_{1}$& -0.015&0.087 &0.097 &0.985&& -0.026     &   0.091  & &0.019& 0.129& &-0.087& 0.087 & &-0.067& 0.084 \\
			&     &   $\beta_{2}$& $<$0.001&0.171 &0.180 &0.975 &&   0.037 &  0.177  & &-0.053& 0.188&  &   0.145& 0.221& &0.055& 0.168  \\
			&1000 &  $\beta_{1}$& -0.004&0.062 &0.065 &0.960 && -0.024     &   0.065   & &0.032& 0.065& &  -0.069& 0.064 & &-0.055& 0.064\\
			&     & $\beta_{2}$& 0.016&0.124 &0.120 &0.955 &&    0.028    &  0.118  & &-0.021& 0.127& &   0.102& 0.137& &0.064& 0.121  \\  			
			3       & 500 &  $\beta_{1}$& 0.011&0.085 &0.095 &0.980&&   0.020   &   0.096  & &-0.007& 0.134&  &-0.043& 0.081& &-0.022& 0.081   \\
			&      &  $\beta_{2}$& -0.004&0.172 &0.178 &0.960 &&   -0.021   &  0.195  & &-0.025& 0.190& &  0.105& 0.221 & &0.024& 0.174  \\
			&1000 &$\beta_{1}$& 0.013&0.062 &0.064 &0.955 &&  0.009    &  0.059  & &0.010& 0.065&  &   -0.026& 0.061& &-0.011& 0.061 \\
			&    &   $\beta_{2}$& -0.002&0.119 &0.121 &0.970&&  -0.009    & 0.123    & &-0.003& 0.124&  & 0.050& 0.129 & &0.014& 0.117 \\
			4        & 500 &$\beta_{1}$& 0.056&0.096 &0.096 &0.930 &&    0.049  &  0.095  & &0.106& 0.115& &  0.001& 0.090 & &0.023& 0.089\\
			&      &  $\beta_{2}$& -0.035&0.150 &0.171 &0.970 &&   0.069   & 0.175   & &-0.079& 0.191& & 0.056& 0.161& &-0.001& 0.143  \\
			&1000 &  $\beta_{1}$& 0.020&0.060 &0.060 &0.959 &&   0.030   &  0.065  & &0.041& 0.063& &  -0.005& 0.054& &0.004& 0.054  \\
			&     &  $\beta_{2}$& -0.013&0.108 &0.108 &0.964 &&    0.048  &  0.122  & &-0.033& 0.119& &  0.025& 0.105& &0.005& 0.102 \\
			5        & 500 & $\beta_{1}$& -0.005&0.093 &0.105 &0.965 &&   0.057   &  0.169  & &0.122& 0.129& & -0.075& 0.088& &-0.058& 0.087 \\
			&   &  $\beta_{2}$& 0.009&0.168 &0.192 &0.985 &&   -0.104   &   0.292 & &-0.119& 0.223&   &    0.135& 0.196 & &0.066& 0.159  \\
			&1000 &$\beta_{1}$& -0.020&0.069 &0.067 &0.943&&   0.018   &   0.090  & &0.045& 0.070& &   -0.080& 0.057 & &-0.076& 0.056\\
			&    & $\beta_{2}$& 0.031&0.138 &0.119 &0.953 &&   -0.027   &  0.158  & &-0.038& 0.132&  & 0.106& 0.109& &0.083& 0.103 \\ 		
			6       & 500 &  $\beta_{1}$&-0.003 & 0.093 & 0.103 & 0.944&&   0.052   &  0.161  & &0.019& 0.116&  &-0.111& 0.080& &-0.092 & 0.080 \\
			&   &   $\beta_{2}$&  0.012& 0.165 & 0.180 & 0.949&&   -0.100   &   0.277 & &-0.008& 0.203 &   &    0.169& 0.194 & &0.095& 0.154  \\
			&1000 &  $\beta_{1}$& -0.035& 0.060 & 0.064 &0.927 &&   0.012   &   0.099 & &-0.058& 0.059 & &   -0.123& 0.049& &-0.115& 0.049 \\
			&    &   $\beta_{2}$& 0.037 & 0.107 &0.111  &0.938 &&   -0.053   & 0.165   & &0.058& 0.121&  &  0.144& 0.109& &0.118& 0.100 \\ 		
			
		\end{tabular}
		\begin{tablenotes}
			\footnotesize
			\item[*]Note:  ``Proposed method'' denotes the proposed deep regression approach,   ``Deep PH '' refers to the deep learning approach  for the partially linear PH model \citep{du2024deep},  ``Spline-Trans'' refers to the sieve  MLE approach for the partially linear additive transformation model   \citep{Yuan2024SIM}, ``Penalized PH'' refers to penalized MLE method for the PH model with linear covariate effects  \citep{zhao2020simultaneous},     ``NPMLE-Trans'' refers to the nonparametric MLE method for the  transformation model   with linear   covariate effects   \citep{zeng2016maximum}.  \\
		\end{tablenotes}
		\end{threeparttable}
	}
	\end{center}
\end{table}

\begin{table}
	\caption{Simulation results for
		the relative error (RE) of  $\hat{\phi}_{\balpha}$ over  $\phi$ and the mean squared  error (MSE) of the estimated survival function.
		Values of  MSE in the table are multiplied by  100.}\label{table2}
	
\begin{center}
	\scalebox{0.8}{
	\begin{threeparttable}
	\begin{tabular}{rrrrrrrrrrrrrrrr}
	&        &        \multicolumn{2}{c} {Proposed method}    & & \multicolumn{2}{c}{Deep PH}    & & \multicolumn{2}{c}{Spline-Trans} & & \multicolumn{2}{c}{ Penalized PH}& & \multicolumn{2}{c}{NPMLE-Trans }\\
	\cline{3-4} \cline{6-7}\cline{9-10}\cline{12-13}\cline{15-16}
	Case   & $n$ &   RE&MSE && RE&MSE&& RE&MSE& & RE&MSE & & RE&MSE \\
	\hline
	1        & 500  & 0.314&0.267 && 0.405    & 0.511  & &1.739& 0.984&&0.294 &0.254 & &0.236& 0.510 \\
	&1000 & 0.230&0.131 &&  0.309   & 0.349  & &1.824& 0.323&&0.196 &0.112 & &0.170&  0.229 \\
	2        & 500 & 0.515&0.739&& 0.640 & 1.226 & &1.037& 0.790&&0.855 &1.679 & &0.854& 1.879  \\
	&1000 & 0.423&0.477 &&   0.518  & 0.873  & &0.740& 0.309&&0.837 &1.558 & &0.841& 1.686  \\   			
	3       & 500 & 0.728&0.545 &&  0.769   & 0.725  & &1.812&2.013&&0.985 &0.881 & &1.012& 1.160  \\
	&1000 & 0.571&0.314 && 0.599   & 0.481  & &1.516& 1.383 &&0.975 &0.797 & &0.985& 0.927  \\
	4        & 500 & 0.477&0.625 &&  0.516   & 1.163 & &2.044& 1.389 &&0.365 &0.412 & &0.320& 0.718  \\
	&1000 & 0.313&0.285 && 0.421    &  1.039 & &1.947& 0.545&&0.263 &0.206 & &0.216& 0.337  \\
	5        & 500 & 0.358&1.606 &&   0.545  & 3.212  & &1.574& 0.900&&0.430 &1.803 & &0.426& 2.489 \\
	&1000 & 0.292&0.984 &&   0.421  &  2.515 & &1.407& 0.369&&0.415 &1.573 & &0.415& 1.889  \\
	6       & 500 & 0.556 & 1.883 &&   0.749  &  3.365 & &1.126& 3.903&&0.767 &2.767 & &0.767& 3.168  \\
	&1000 & 0.488 & 1.262 &&  0.630   &  2.603 & &1.078& 3.199&&0.761 &2.545 & &0.763& 2.740  \\  	
				
	\end{tabular}
	\begin{tablenotes}
	\footnotesize
	\item[*]Note:  ``Proposed method'' denotes the proposed deep regression approach,   ``Deep PH '' refers to the deep learning approach  for the partially linear PH model \citep{du2024deep},  ``Spline-Trans'' refers to the sieve  MLE approach for the partially linear additive transformation model   \citep{Yuan2024SIM}, ``Penalized PH'' refers to penalized MLE method for the PH model with linear covariate effects  \citep{zhao2020simultaneous},     ``NPMLE-Trans'' refers to the nonparametric MLE method for the  transformation model   with linear   covariate effects   \citep{zeng2016maximum}. \\
		\end{tablenotes}
	\end{threeparttable}
	}
\end{center}
\end{table}

To manifest the usefulness of the proposed deep regression method,
we reanalyzed the above simulated data with   four  comparative  state-of-the-art methods.
They include  the deep learning approach for the partially linear PH model  abbreviated as  ``Deep PH''   \citep{du2024deep}, the sieve  maximum likelihood  estimation (MLE) approach for the partially linear additive transformation model   abbreviated as  ``Spline-Trans''   \citep{Yuan2024SIM},
the penalized MLE method for the PH model with linear covariate effects     abbreviated as  ``Penalized PH''  \citep{zhao2020simultaneous},
and the nonparametric MLE method for the  transformation model   with linear   covariate effects   abbreviated as  ``NPMLE-Trans''    \citep{zeng2016maximum}.
For each $i= 1, \ldots, n$,     the  partially linear additive  transformation model in \cite{Yuan2024SIM}
specifies  the cumulative hazard function of $T_i$ given   $\bX_i$ and $ \bW_i$   as
$
\Lambda(t \mid  \bX_i, \bW_i)=G\{\Lambda(t) \exp \{\bbeta^{\top} \bX_i+\sum_{k=1}^{K}  \phi_k(W_{ik}) \}\},
$
where $\bbeta$ is a vector of unknown regression coefficients, $W_{ik}$ is the $k$th component of $\bW_i$,
and $\phi_k$ is an unknown function with $k=1, \ldots, K.$
In \cite{Yuan2024SIM}'s approach,  $B$-splines  were used to estimate  each $\phi_k(W_{ik})$.

The results obtained from the above  four  methods on the regression parameter estimates are also summarized in Table \ref{table1}.
In the case of linear covariate effects (e.g., Case 1 and Case 4),     ``Penalized PH''  and  ``NPMLE-Trans''    perform satisfactorily and are comparable to the proposed method.
This finding is anticipated since the model used in ``Penalized PH''  and  ``NPMLE-Trans'' is correctly specified under the case of linear covariate effects.
``Spline-Trans''  yields significant estimation biases in Case 4 when the sample size is 500, partially because  ``Spline-Trans''  suffers from the curse of dimensionality if the length of $\bW$ is considerable.
Regarding the settings of nonlinear covariate effects,  ``Penalized PH''  and  ``NPMLE-Trans''  generally exhibit large estimation biases since the assumed models are misspecified.
Although  ``Spline-Trans''  is applicable to estimate a partially linear  model, it can only  accommodate a few nonlinear covariates in an additive form as suggested by \cite{Yuan2024SIM}.
This fact and the curse of dimensionality restrict the empirical performance   of ``Spline-Trans''  on   estimating  $\bbeta$ sometimes (e.g., Cases 5 and 6).
``Deep PH''  exhibits   considerable estimation biases in the settings of  $n=500$ and 10 nonlinear covariates (i.e., Cases 5 and  6). Such inferior performance arises partially because the estimate of cumulative hazard function cannot remain monotone
with the optimization procedure of ``Deep PH''.

To evaluate the prediction accuracy of the proposed deep regression method,
we used  the relative error (RE) of  $\hat{\phi}$ over  $\phi$ and the mean squared  error (MSE) of the estimated survival function as in \cite{zhong2022deep} and
\cite{sun2023neural}.
In particular,
RE is defined  as
$$
\operatorname{RE}(\hat{\phi})=\left[\frac{\frac{1}{n_1} \sum_{i=1}^{n_1}\left\{ \hat{\phi}\left(\bW_i\right) -\phi\left(\bW_i\right)\right\}^2}{\frac{1}{n_1} \sum_{i=1}^{n_1}\left\{\phi\left(\bW_i\right)\right\}^2}\right]^{1 / 2},
$$
and MSE of   the estimated survival function is defined as
\begin{flalign}
	\operatorname{MSE}(\widehat{S})=\frac{1}{n_1} \sum_{i=1}^{n_1} \frac{1}{T_i} \int_0^{T_i}\left\{S\left(t \mid \bX_i, \bW_i\right)-\hat{S}\left(t \mid \bX_i, \bW_i\right)\right\}^2 d t,
	\label{LS}
\end{flalign}
where $S(t \mid \bX_i, \bW_i) =  \exp ( - G[ \Lambda (t) \exp\{  \bbeta ^{\top}  \bX_i + \phi(\bW_i)\} ])$,  $\hat{S}(t \mid \bX_i, \bW_i) =  \exp ( -  G[\hat{\Lambda} (t)  \\  \exp  \{  \hat{\bbeta} ^{\top}  \bX_i + \hat{\phi}(\bW_i)\}] ),$
and   $\left\{T_i, \bX_i, \bW_i; i=1, \ldots, n_1\right\}$ is   the test dataset with $n_1 = 0.2 n$.
Smaller values of REs shown in Table \ref{table2} indicate a better prediction performance of the proposed method over the  ``Deep PH''  and   ``Spline-Trans''.
``Deep PH''  generally produces larger  MSEs than the proposed method,
especially under  Cases 4--6.
In addition, it can be observed that
``Spline-Trans'' can only yield comparable results to the proposed method under the additive models (e.g., Case 2 and Case 5) when the sample size is large. 
In more complex settings (e.g., Case 3 and Case 6), the proposed deep approach consistently  outperforms the ``Spline-Trans''.
As for  ``Penalized PH'' and ``NPMLE-Trans'',  they both yield conspicuously large REs and MSEs except for the cases of linear covariate effects (i.e., Case 1 and Case 4). The above comparison results all manifest the significant advantages of the proposed method in terms of prediction accuracy.

In Section S.3 of the supplementary materials, we conducted simulations under other regression models, such as  the PO model.
The  results given in Tables S1 and S2  suggest similar conclusions as  mentioned above
regarding the comparisons of the proposed method,
``Spline-Trans''  and ``NPMLE-Trans''.
The performances of ``deep PH" and ``penalized PH" deteriorate remarkably  due to using a misspecified PH model. This phenomenon further demonstrates the practical utility of developing a general deep regression model.

\section{Application}
\label{sec:appli}
We applied the proposed deep regression method to a set of interval-censored failure time data from the Alzheimer's Disease Neuroimaging Initiative (ADNI) study initiated in 2004.
This ADNI study recruited over 2,000 participants from   North America, documenting their cognitive conditions categorized into three groups: cognitively normal, mild cognitive impairment (MCI), and Alzheimer's disease (AD).
In our analysis, we followed \cite{li2017prediction}  and other researchers by concentrating  on individuals  who had experienced cognitive decline in the MCI group.
Our primary objective is to investigate the crucial risk factors for  the progression of mild cognitive impairment to AD,
thereby  enhancing individuals' quality of life and extending their cognitive health span.
The failure time of interest was defined as the duration from participant recruitment to the AD conversion.
Since  each participant was examined intermittently for the AD development,
the  AD occurrence time  cannot be attained precisely and suffers from   general interval censoring.
In other words,  the exact  AD onset  time of  each participant
is only known to fall into a time interval.
The dataset consists  of a total of 938 participants with 331  interval-censored observations and 579  right censored ones.

We focused on 14 potential risk factors for AD, consisting of a genetic covariate:
\texttt{APOE$\epsilon$4} (coded as 0, 1, or 2 based on the number of APOE 4 alleles),
four demographic covariates:
\texttt{Age} (given in years),
\texttt{Gender} (1 for male and 0 for female),
\texttt{Education} (years of education),
\texttt{Marry} (for married and 0 otherwise),
and nine test scores,  which are explained in Section S.4 of the supplementary materials due to the space restriction.
We randomly divided the entire data into a training dataset consisting of 742 individuals and a test dataset of 196 observations,
where the latter was used to evaluate the prediction performance and select the optimal hyperparameters as well as the optimal transformation function of the proposed method.

Note that  \texttt{APOE$\epsilon$4}, \texttt{Age} and   \texttt{Gender} were reported as  crucial   factors for developing AD in many   studies,  including
\cite{Sun2021SEM}, \cite{Mielke2022} and
\cite{Livingston2023}.
Our analysis attempted to investigate these three potential risk factors while controlling for other variables. Therefore, we included \texttt{APOE$\epsilon$4}, \texttt{Age} and  \texttt{Gender} in the linear term of model \eqref{model}, aiming to examine their effects quantitatively on the progression of AD in individuals with mild cognitive impairment. Other covariates 
were treated as control variables adjusted in the regression model.
That is, we assumed that the failure time of interest followed the partially linear transformation model with the conditional hazard function:
$$\Lambda(t \mid \bX, \bW) =G[ \Lambda(t) \exp\{X_1\beta_1 +X_2\beta_2 +X_3\beta_3 + \phi(\bW)\}],$$
where $G(x) = \log(1 + rx)/r$ with $r \geq 0$, $X_1=$  \texttt{Age},   $X_2=$ \texttt{Gender},  $X_3=$ \texttt{APOE$\epsilon$4} and
$\bW$ includes all the aforementioned covariates except $X_1$ ,$X_2$ and  $X_3$.
Note that the above transformation model with $r = 0$ (i.e., $G(x)=x$) corresponds to the partially linear PH model while setting $r = 1$ in the above model yields the partially linear  PO model.

We used cubic splines to approximate $\Lambda(\cdot)$ and considered a sequence of integers ranging from 1 to 8 to determine the optimal number of interior knots. These interior knots were placed equally spaced quantiles between the minimum and maximum observation times. For the transformation function $G(\cdot)$, we let $r$ vary from  0 to 5 with an increment of 0.1. In constructing the neural network, as in the simulation studies,  we considered 2 or 3 hidden layers, 50 neurons in each hidden layer, and the SeLU activation function. We set the tuning parameter in the $L_1$ penalty to  0.1, 0.05 or  0.01 and the learning rate to $10^{-3}$ or $5 \times 10^{-3}$ to train the constructed network. The mini-batch size in the SGD algorithm was set to 30. The neural network was trained for 20 epochs during each iteration of the proposed EM algorithm. Meanwhile, we used the dropout regularization with a rate of 0.1 to prevent overfitting. To measure the prediction performance of the proposed method, we utilized an integrated Brier score (IBS)  calculated on the test data.
The explicit definition of IBS can be found  in Section S.4 of the supplementary materials.  A smaller value of IBS corresponds to a better prediction performance. The analysis showed that the partially linear  model with $r=2.6$ was optimal.

Table \ref{tab5} presents the   analysis results for   parametric covariate effects with the proposed deep
and   four  comparative methods described in the simulations, including ``Deep PH ''  \citep{du2024deep},``Penalized PH'' \citep{zhao2020simultaneous},  ``Spline-Trans'' \citep{Yuan2024SIM},  and ``NPMLE-Trans''  \citep{zeng2016maximum}.
Results include  the  regression parameter estimates (EST),  the estimated standard errors (SE), and   $p$-values.
Specifically,  the SEs of  the proposed deep method and ``NPMLE-Trans''   were obtained via the profile likelihood method,
while the SEs of  ``Penalized PH'' and ``Spline-Trans''   were calculated using nonparametric bootstrapping with a bootstrap size 100.
The results obtained from the proposed deep learning method with $r=2.6$, as presented in Table \ref{tab5},   reveal that  males  exhibit a significantly higher risk of developing   AD  compared to females. The presence of the \texttt{APOE$\epsilon$4}  allele demonstrates a notably positive association with the risk of AD, suggesting that carrying the \texttt{APOE$\epsilon$4}  allele increases the hazard of developing AD.
In particular, individuals carrying two \texttt{APOE$\epsilon$4}  alleles are at an even greater risk than those carrying only one allele.
Furthermore, among individuals  who had already experienced cognitive decline,  \texttt{Age} is also   a significant risk factor for AD by our method; being older
represents a  higher risk of developing AD.
The proposed  method with $r=0$ (PH model) or 1 (PO model)
leads to the same conclusions as the optimal model.
Regarding the results of the four comparative methods, compared to the proposed method, the most notable difference
is that the comparative methods all recognized  \texttt{Age} as insignificant.
This conclusion may be erroneous since  age has been  widely acknowledged
as one of the most significant influential factors for AD  \citep[e.g.,][]{James2019, Livingston2023}.


\begin{table}
\caption{Analysis results of the ADNI data.
	``EST''  and ``SE'' denote the regression parameter estimate and its standard error estimate, respectively.}\label{tab5}

\begin{center}
	\scalebox{0.7}{
	\begin{threeparttable}
	\begin{tabular}{lrrrrrrrrrrr}
	 & \multicolumn{3}{c} {\texttt{Age}}        & & \multicolumn{3}{c}{ \texttt{Gender}} & & \multicolumn{3}{c}{ \texttt{APOE$\epsilon$4}}  \\
	\cline{2-4} \cline{6-8} \cline{10-12}

	Method      & EST  & SE     & $p$-value              &    & EST        & SE & $p$-value  &    & EST        & SE & $p$-value   \\
	\hline
	Proposed method ($r$=2.6) & 0.089 & 0.029 & 0.002        &     & 0.434       &0.045     & $<0.001$ &     & 0.753      &0.045     &  $<0.001$   \\
	Proposed method ($r$=0)    & 0.092 & 0.029 & 0.002          &     & 0.378       &0.048    &  $<0.001$ &     & 0.374      &0.045     &  $<0.001$    \\
	Proposed method ($r$=1)    & 0.108 & 0.032 & 0.001          &     & 0.464       &0.034    &  $<0.001$ &     & 0.568      &0.034     &  $<0.001$    \\
	Deep PH                              & 0.067 & 0.111 & 0.543          &     & 0.384       &0.172    &  0.025 &     & 0.319      &0.114     &  0.005   \\
	Penalized PH                        & 0 &0.079 & 1.000             &      &0.320        & 0.199    & 0.108&      &0.317      & 0.124  & 0.011   \\
	Spline-Trans  ($r$=2.6)       & -0.089 & 0.169  & 0.600   &      &0.761       & 0.350   & 0.030  &      &0.742      & 0.264 & 0.005     \\
	Spline-Trans  ($r$=0)           & 0.342 & 0.403  & 0.396   &      &0.336        & 0.469    & 0.474  &      &1.229        & 0.402    & 0.002    \\
	Spline-Trans  ($r$=1)      & 0.040 & 0.141 & 0.777          &     & 0.596       &0.286    &  0.037 &     & 0.290      &0.187     &  0.121    \\
	NPMLE-Trans ($r$=2.6)   & 0.062 &0.121  & 0.609             &      &0.524        & 0.241& 0.030 &      &0.671        & 0.164    & $<0.001$       \\
	NPMLE-Trans ($r$=0)   & 0.123 &0.073  & 0.092             &      &0.422        & 0.145    & 0.004 &      &0.388      & 0.093    & $<0.001$       \\
	NPMLE-Trans ($r$=1)    & 0.106 & 0.096 & 0.270          &     & 0.482       &0.198    &  0.015 &     & 0.536      &0.129    &  $<0.001$   \\
\end{tabular}
\begin{tablenotes}
\footnotesize
\item[*]Note:  ``Proposed method'' denotes the proposed deep regression approach,   ``Deep PH '' refers to the deep learning approach  for the partially linear PH model \citep{du2024deep},  ``Spline-Trans'' refers to the sieve  MLE approach for the partially linear additive transformation model   \citep{Yuan2024SIM}, ``Penalized PH'' refers to penalized MLE method for the PH model with linear covariate effects  \citep{zhao2020simultaneous},     ``NPMLE-Trans'' refers to the nonparametric MLE method for the  transformation model   with linear   covariate effects   \citep{zeng2016maximum}.
$r = 0$ and $r = 1$  correspond  to the  PH and  PO models, respectively.\\
\end{tablenotes}
\end{threeparttable}
		}
	\end{center}
\end{table}

In Section S.4 of the supplementary materials,  
we conducted a 5-fold cross-validation
to investigate the prediction performance of the proposed deep method.
Figure S1 presents the obtained IBS values across all folds under the proposed deep regression method with the optimal model ($r=2.6$)
and the above comparative  methods.
It shows that the proposed method leads to the smallest IBS value except for the first fold,
confirming our method's powerful prediction ability.

\section{Discussion and Concluding Remarks}
\label{sec:conc}
Inspired by DNN's powerful approximation ability, this study proposed a novel deep regression approach for analyzing interval-censored data with a class of partially linear transformation models. Our proposed method leveraged the spirit of sieve estimation, in which the   nonparametric component of the   model was approximated by DNN.
Since incorporating deep learning into time-to-event analysis is relatively new, it is easy to envision some potential future directions.
For example, in practical applications, the failure times of interest are often accompanied by complex features (e.g.,  left truncation and competing risks  \citep{shen2009analyzing, MaoLin2017}
or garnered from intractable sampling strategies, such as group testing and case-cohort design  \citep{Zhou2017Bka, Li2024Bka}. Thus, generalizations of the proposed deep regression approach to account for the settings above warrant future research.
Furthermore, multivariate interval-censored data arise frequently in many real-life studies, where each individual may undergo multiple events of interest that suffer from interval censoring \citep{Zeng2017Bka, Sun2022Bio}. Naturally, the multiple event times of the same individual are correlated, which poses considerable challenges in parameter estimation and theoretical development. Therefore, meaningful future work is to explore some deep learning methods tailored to multivariate interval-censored data.

\section*{Acknowledgment}

This research was partly supported by the National Natural Science Foundation of China (12471251), and
the Research Grant Council of the Hong Kong Special Administrative Region
(14301918, 14302519).

\bigskip
%
%
%
%
%

%
%

\begin{thebibliography}{}
\bibitem[Bauer and Kohler (2019)]{bauer2019deep}
Bauer, B., Kohler, M.(2019).
On deep learning as a remedy for the curse of dimensionality in nonparametric regression.
{\em The Annals of Statistics}~{\bf 47}(4), 2261 -- 2285.




 
\bibitem[Chen et al.(2020)]{chen2020arelu}
Chen, D., Li, J., Xu, K.(2020).
Arelu: Attention-based rectified linear unit.
{\em arXiv:  2006.13858}.

\bibitem[Chen and Jin (2006)]{ChenJin2006}
Chen, K., Jin, Z.(2006).
Partial Linear Regression Models for Clustered Data.
{\em Journal of the American Statistical Association}~{\bf 101}(473), 195--204.
 
\bibitem[Cybenko(1989)]{cybenko1989approximation}
Cybenko, G.(1989).
Approximation by superpositions of a sigmoidal function.
{\em  Mathematics of Control, Signals and Systems}~{\bf 2},303--314.

\bibitem[Devlin et al.(2019)]{devlin2018bert}
Devlin, J., Chang, M.W., Lee, K., Toutanova, K.(2019).
BERT: Pre-training of Deep Bidirectional Transformers for Language Understanding.
{\em arXiv: 1810.04805}.

 
\bibitem[Du et al.(2024)]{du2024deep}
Du, M., Wu, Q., Tong, X., Zhao, X.(2024).
Deep learning for regression analysis of interval-censored data.
{\em Electronic Journal of Statistics}~{\bf 18}(2),4292--4321.

\bibitem[Dubey et al.(2022)]{dubey2022activation}
Dubey, S.R., Singh, S.K., Chaudhuri, B.B.(2022).
Activation functions in deep learning: A comprehensive survey and benchmark.
{\em Neurocomputing}~{\bf 503},92--108.


\bibitem[Fan et al.(1997)]{FanCox1997}
Fan, J., Gijbels, I., King, M. (1997).
Local likelihood and local partial likelihood in hazard regression.
{\em The Annals of Statistics}~{\bf 25}(4), 1661--1690.
	
\bibitem[Faraggi and Simon(1995)]{faraggi1995neural}
Faraggi, D., Simon, R. (1995).
A neural network model for survival data.
{\em Statistics in Medicine}~{\bf 14}(1), 73--82.
 
\bibitem[Glorot and Bengio (2010)]{glorot2010understanding}
Glorot, X., Bengio, Y. (2010).
Understanding the difficulty of training deep feedforward neural networks.
{\em Proceedings of the 13th International Conference on Artificial Intelligence and Statistics}~{\bf 9}, 249--256.
 
\bibitem[Goodfellow et al.(2016)]{bengio2017deep}
Goodfellow, I., Bengio, Y., Courville, A. (2016).
{\em Deep learning}, 
MIT press.	

\bibitem[He et al.(2015)]{he2015delving}
He, K., Zhang, X., Ren, S., Sun, J. (2015).
Delving deep into rectifiers: Surpassing human-level performance on imagenet classification.
{\em 2015 IEEE International Conference on Computer Vision}~{\bf 9},1026--1034.

\bibitem[Hornik et al.(1989)]{hornik1989multilayer}
Hornik, K., Stinchcombe, M., White, H. (1989).
Multilayer feedforward networks are universal approximators.
{\em Neural Networks}~{\bf 2}(5), 359--366.
 
\bibitem[Huang(1996)]{huang1996efficient}
Huang, J. (1996).
Efficient estimation for the proportional hazards model with interval censoring.
{\em The Annals of Statistics}~{\bf 24}(2), 540--568.

\bibitem[Huang(1999)]{Huang1999}
Huang, J. (1999).
Efficient Estimation of the Partly Linear Additive {C}ox Model.
{\em The Annals of Statistics}~{\bf 27}(5), 1536--1563.
 
\bibitem[James et al.(2019)]{James2019}
James, S.L., Theadom, A., Ellenbogen, R.G., et al. (2019).
Global, regional, and national burden of traumatic brain injury and spinal cord injury, 1990-2016: a systematic analysis for the Global Burden of Disease Study 2016.
{\em Lancet Neurology}~{\bf 18}(1), 56--87.

 
\bibitem[Katzman et al.(2018)]{katzman2018deepsurv}
Katzman, J.L., Shaham, U., Cloninger, A., Bates, J., Jiang, T., Kluger, Y. (2018).
DeepSurv: personalized treatment recommender system using a Cox proportional hazards deep neural network.
{\em BMC Medical Research Methodology}~{\bf 18}, 24.
 
\bibitem[Klambauer et al.(2017)]{klambauer2017self}
Klambauer, G., Unterthiner, T., Mayr, A., Hochreiter, S. (2017).
Self-Normalizing Neural Networks.
{\em arXiv: 1706.02515}.

\bibitem[Kosorok et al.(2004)]{kosorok2004robust}
Kosorok, M.R., Lee, B.L., Fine, J.P. (2004).
Robust inference for univariate proportional hazards frailty regression models.
{\em The Annals of Statistics}~{\bf 32}(10), 1448--1491.
 
 

 
\bibitem[Kvamme et al.(2019)]{kvamme2019time}
Kvamme, H., Borgan, O., Scheel, I. (2019).
Time-to-event prediction with neural networks and Cox regression.
{\em Journal of Machine Learning Research}~{\bf 20}(129), 1--30.

 

 
\bibitem[Lee et al.(2022)]{LeeCY2022}
Lee, C.Y., Wong, K.Y., Lam, K.F., Xu, J. (2022).
Analysis of clustered interval-censored data using a class of semiparametric partly linear frailty transformation models.
{\em Biometrics}~{\bf 78}(1),  165--178.

 

 
\bibitem[Li et al.(2017)]{li2017prediction}
Li, K., Chan, W., Doody, R.S., Quinn, J., Luo, S., Initiative, A.D.N., et al. (2017).
Prediction of conversion to Alzheimer’s disease with longitudinal measures and time-to-event data.
{\em Journal of Alzheimer's Disease}~{\bf 58}(2), 361--371.

 
\bibitem[Li et al.(2024)]{Li2024Bka}
Li, S., Hu, T., Wang, L., McMahan, C.S., Tebbs, J.M. (2024).
Regression analysis of group-tested current status data.
{\em Biometrika}~{\bf 111}(3), 1047--1061.

 
\bibitem[Li and Peng (2023)]{LiPeng2023}
Li, S., Peng, L. (2023).
Instrumental variable estimation of complier causal treatment effect with interval-censored data.
{\em Biometrics}~{\bf 79}(1), 253--263.

 
\bibitem[Livingston et al.(2023)]{Livingston2023}
Livingston, G., Huntley, J., Sommerlad, A. (2023).
Dementia prevention, intervention, and care: 2020 report of the Lancet Commission.
{\em Lancet}~{\bf 402}(10408), 1132--1132.

 
\bibitem[Lu and McMahan(2018)]{lu2018partially}
Lu, M., McMahan, C.S. (2018).
A partially linear proportional hazards model for current status data.
{\em Biometrics}~{\bf 74}(4), 1240--1249.

 
\bibitem[Lu et al.(2007)]{lu2007estimation}
Lu, M., Zhang, Y., Huang, J. (2007).
Estimation of the Mean Function with Panel Count Data Using Monotone Polynomial Splines.
{\em Biometrika}~{\bf 94}(3), 705--718.

 
\bibitem[Lu and Zhang(2010)]{lu2010estimation}
Lu, W., Zhang, H.H. (2010).
On estimation of partially linear transformation models.
{\em Journal of the American Statistical Association}~{\bf 105}(490), 683--691.

 
\bibitem[Mao and Lin (2017)]{MaoLin2017}
Mao, L., Lin, D.Y. (2017).
Efficient estimation of semiparametric transformation models for the cumulative incidence of competing risks.
{\em Journal of the Royal Statistical Society: Series B}~{\bf 79}(2), 573--587.

 
\bibitem[Mielke et al.(2022)]{Mielke2022}
Mielke, M.M., Aggarwal, N.T., Vila-Castelar, C., et al.(2022).
Consideration of sex and gender in Alzheimer's disease and related disorders from a global perspective.
{\em  Alzheimer's \& Dementia}~{\bf 18}(2), 1--18.

 

 
\bibitem[Ramsay(1988)]{Ramsay1988}
Ramsay, J.O. (1988).
Monotone regression splines in action.
{\em Statistical Science}~{\bf 32}, 425--441.
	

\bibitem[Ren et al.(2019)]{ren2019deep}
Ren, K., Qin, J., Zheng, L., Yang, Z., Zhang, W., Qiu, L., Yu, Y. (2019).
Deep recurrent survival analysis.
{\em AAAI Conference on Artificial Intelligence}~{\bf 33}(01), 4798--4805.

 
\bibitem[Ren et al.(2017)]{ren2016faster}
Ren, S., He, K., Girshick, R., Sun, J. (2017).
Faster R-CNN: Towards real-time object detection with region proposal networks.
{\em IEEE Transactions on Pattern Analysis and Machine Intelligence}~{\bf 39}(6), 1137--1149.

 
\bibitem[Schmidt-Hieber(2020)]{schmidt2020nonparametric}
Schmidt-Hieber, J. (2020).
Nonparametric regression using deep neural networks with ReLU activation function.
{\em The Annals of Statistics}~{\bf 48}(4), 1875 -- 1897.

 
\bibitem[Shen et al.(2009)]{shen2009analyzing}
Shen, Y., Ning, J., Qin, J. (2009).
Analyzing length-biased data with semiparametric transformation and accelerated failure time models.
{\em Journal of the American Statistical Association}~{\bf 104}(487), 1192--1202.

 
\bibitem[Silver et al.(2017)]{silver2017mastering}
Silver, D., Schrittwieser, J., Simonyan, K., Antonoglou, I., Huang, A., Guez, A., Hubert,
T., Baker, L., Lai, M., Bolton, A., et al. (2017).
Mastering the game of go without human knowledge.
{\em Nature}~{\bf 550}(7676), 354--359.

 
\bibitem[Srivastava et al.(2014)]{srivastava2014dropout}
Srivastava, N., Hinton, G., Krizhevsky, A., Sutskever, I., Salakhutdinov, R. (2014).
Dropout: a simple way to prevent neural networks from overfitting.
{\em Journal of Machine Learning Research}~{\bf 15}(1), 1929--1958.

 
\bibitem[Sun(2006)]{sun2006statistical}
Sun, J. (2006).
{\em The Statistical Analysis of Interval-Censored Failure Time Data}, 
Springer, New York.
 
\bibitem[Sun et al.(2022)]{Sun2022Bio}
Sun, L., Li, S., Wang, L., Song, X., Sui, X. (2022).
Simultaneous variable selection in regression analysis of multivariate interval-censored data.
{\em Biometrics}~{\bf 78}(4), 1402--1413.

 
\bibitem[Sun et al.(2021)]{Sun2021SEM}
Sun, R., Zhou, X., Song, X. (2021).
Bayesian Causal Mediation Analysis with Latent Mediators and Survival Outcome.
{\em Structural Equation Modeling: A Multidisciplinary Journal}~{\bf 28}(5), 778--790.

 
\bibitem[Sun and Ding(2023)]{sun2023neural}
Sun, T., Ding, Y. (2023).
Neural network on interval-censored data with application to the prediction of Alzheimer's disease.
{\em Biometrics}~{\bf 79}(3), 2677--2690.

 
\bibitem[Sun et al.(2024)]{sun2024penalized}
Sun, Y., Kang, J., Haridas, C., Mayne, N., Potter, A., Yang, C.F., Christiani, D.C., Li,Y. (2024).
Penalized deep partially linear cox models with application to CT scans of lung cancer patients.
{\em Biometrics}~{\bf 80}(1), ujad024.

 

 
\bibitem[Telgarsky(2015)]{telgarsky2015representation}
Telgarsky, M. (2015).
Representation benefits of deep feedforward networks.
{\em arXiv: 1509.08101}.

 
\bibitem[Tong and Zhao(2022)]{tong2022deep}
Tong, J., Zhao, X. (2022).
Deep survival algorithm based on nuclear norm.
{\em Journal of Statistical Computation and Simulation}~{\bf 92}(9), 1964--1976.

 

 
\bibitem[Wang et al.(2016)]{wang2016flexible}
Wang, L., McMahan, C.S., Hudgens, M.G., Qureshi, Z.P. (2016).
A flexible, computationally efficient method for fitting the proportional hazards model to interval-censored data.
{\em Biometrics}~{\bf 72}(1), 222--231.

 
\bibitem[Wu et al.(2024)]{wu2024deep}
Wu, Q., Tong, X., Zhao, X. (2024).
Deep partially linear cox model for current status data.
{\em Biometrics}~{\bf 80}(2), ujae024.

 
\bibitem[Xie and Yu (2021)]{Xie2021SIM}
Xie, Y., Yu, Z. (2021).
Promotion time cure rate model with a neural network estimated nonparametric component.
{\em Statistics in Medicine}~{\bf 40}(15),3516--3532.

 
\bibitem[Yuan et al.(2024)]{Yuan2024SIM}
Yuan, C., Zhao, S., Li, S., Song, X. (2024).
Sieve Maximum Likelihood Estimation of Partially Linear Transformation Models With Interval-Censored Data.
{\em Statistics in Medicine}~{\bf 43}(30), 5765 -- 5776.

\bibitem[Zeng et al.(2017)]{Zeng2017Bka}
Zeng, D., Gao, F., Lin, D.Y. (2017).
Maximum likelihood estimation for semiparametric regression models with multivariate interval-censored data.
{\em Biometrika}~{\bf 104}(3), 505--525.

 
\bibitem[Zeng and Lin(2006)]{zeng2006}
Zeng, D., Lin, D.Y. (2006).
Efficient estimation of semiparametric transformation models for counting processes.
{\em Biometrika}~{\bf 93}(3), 627--640.

 
\bibitem[Zeng et al.(2016)]{zeng2016maximum}
Zeng, D., Mao, L., Lin, D. (2016).
Maximum likelihood estimation for semiparametric transformation models with interval-censored data.
{\em Biometrika}~{\bf 103}(2), 253--271.

 
\bibitem[Zhang et al.(2010)]{zhang2010spline}
Zhang, Y., Hua, L., Huang, J. (2010).
A spline-based semiparametric maximum likelihood estimation method for the {C}ox model with interval-censored data.
{\em Scandinavian Journal of Statistics}~{\bf 37}(2), 338--354.

 
\bibitem[Zhao et al.(2020)]{zhao2020simultaneous}
Zhao, H., Wu, Q., Li, G., Sun, J. (2020).
Simultaneous estimation and variable selection for interval-censored data with broken adaptive ridge regression.
{\em Journal of the American Statistical Association}~{\bf 115}(529), 204-216.

 
\bibitem[Zhong et al.(2022)]{zhong2022deep}
Zhong, Q., Mueller, J., Wang, J.L. (2022).
Deep learning for the partially linear Cox model.
{\em The Annals of Statistics}~{\bf 50}(3), 1348--1375.

 

 
\bibitem[Zhou et al.(2017)]{Zhou2017Bka}
Zhou, Q., Zhou, H., Cai, J. (2017).
Case-cohort studies with interval-censored failure time data.
{\em Biometrika}~{\bf 104}(1), 17--29.

 

\end{thebibliography}


\end{document}


\def\spacingset#1{\renewcommand{\baselinestretch}%
{#1}\small\normalsize} \spacingset{1}


\if0\blind
{
\date{}
  \title{\bf Supplementary Materials for ``Interpretable  Deep Regression Models with Interval-Censored Failure Time Data''}
  \author{Changhui Yuan\\
  	School of Mathematics, Jilin University\\
  	and \\
  	Shishun Zhao \\
  	School of Mathematics, Jilin University\\
  	and \\
  	Shuwei Li  \\
  	School of Economics and Statistics, Guangzhou University\\
  	and \\
  	Xinyuan Song \\
  	Department of
  	Statistics, Chinese University of Hong Kong\\
  	and \\
  	Zhao Chen  \\
  	School of Data Science, Fudan University}
  \maketitle
} \fi

\if1\blind
{
  \bigskip
  \bigskip
  \bigskip
  \begin{center}
    {\LARGE\bf  Supplementary Materials for ``Interpretable  Deep Regression Models with Interval-Censored Failure Time Data''}
\end{center}
  \medskip
} \fi

\bigskip

\spacingset{1.9} 
\section*{Section S.1:  Estimation  Procedure}

We propose an EM algorithm to conduct the sieve maximum likelihood estimation based on likelihood (4) of the main manuscript.
We first introduce a three-stage data augmentation procedure to construct the complete data likelihood with a tractable form.
By considering the class of frailty-induced logarithmic transformations introduced in Section 2.1  of the main manuscript,
the observed data likelihood  (4)  of the main manuscript can be equivalently formulated as
\begin{equation}\tag{S1}\label{obsLik-3}
	\begin{aligned}
		\mathcal{L} _ {1} &(\bbeta,  \bgamma,\balpha) \\
		&=  \prod _ { i = 1 } ^ { n } \int _ { 0 } ^ { \infty }\{ 1 - \exp  \{ -  \Lambda_ {\bgamma} ( R_i ) \exp  \{ \bbeta ^ {\top}  \bX_i +\phi_{\balpha} (\bW_i)\} \eta _ { i }\}\} ^ { \delta _ { L,i } }\\
		&\times \left(\exp  \{ -  \Lambda_ {\bgamma} (L_i) \exp  \{ \bbeta ^ {\top}  \bX_i + \phi_{\balpha} ( \bW_i)\} \eta _ { i }\} -
		\exp  \{ -  \Lambda_ {\bgamma} (R_i) \exp  \{ \bbeta ^ {\top}  \bX_i + \phi_{\balpha} (\bW_i)\} \eta _ { i }\} \right) ^ { \delta _ {I,i } }\\
		&\times \exp  \{ -  \Lambda_ {\bgamma} (L_i ) \exp  \{ \bbeta ^ {\top}  \bX_i + \phi_{\balpha} ( \bW_i)\} \eta _ { i }\} ^ { \delta _ {R,i } } f(\eta_i \mid r) {\rm d} \eta_i.
	\end{aligned}
\end{equation}

\addtolength{\textheight}{.5in}%

For subject $i$,
define $t_{i2} = R_i (\delta_{I,i}=1) + L_i(\delta_{R,i}=1)$ if $\delta_{L,i} = 0$, and
$t_{i1} = R_i (\delta_{L,i}=1) + L_i(\delta_{L,i}=0)$.
We introduce  two independent Poisson random variables: $Z_{i} \sim Pois( \Lambda_ {\bgamma} ( t_{i1} ) $ $\exp  \{ \bbeta ^ {\top}  \bX_i +  \phi_{\balpha} ( \bW_i  )\}\eta _ { i }) $ and $ Y_{ i } \sim Pois(\{\Lambda_ {\bgamma} (t_{i2})-\Lambda_ {\bgamma} (t_{i1})\} \exp  \{ \bbeta ^ {\top}  \bX_i +  \phi_{\balpha}  ( \bW_i )\}\eta _ { i } )$, where $ Pois(\nu) $ indicates the Poisson distribution with mean $ \nu $.
Moreover, by  the spline representation of $\Lambda_ {\bgamma}(t)$, we decompose $Z_i$ and $Y_i$ as $Z_i =  \sum_{l=1}^{L_n}  Z_{il}$ and $Y_i =  \sum_{l=1}^{L_n}  Y_{il}$,
where $Z_{i1}, \ldots, Z_{iL_n}, Y_{i1}, \ldots, Y_{iL_n}$ are independent  Poisson random variables,  the mean of $Z_{il}$ is  $\gamma _ { l } M _ { l } (t_{i1})\exp  \{ \bbeta ^ { \top }  \bX_i +  \phi_{\balpha} ( \bW_i )\}\eta _ { i } $,
and   the  mean of  $Y_{il}$ is  $\gamma _ { l } \{M _ { l } (t_{i2}) - M _ { l } (t_{i1})\}  \exp  \{ \bbeta ^ {\top}  \bX_i +  \phi_{\balpha} ( \bW_i )\}\eta _ { i } $
for $l = 1, \ldots, L_n$.
Then,  the likelihood \eqref{obsLik-3}  can be equivalently expressed as
\begin{equation*}
	\begin{aligned}
		\mathcal{L} _ {2} (\bbeta, \bgamma,\balpha) = \prod _ { i = 1 } ^ { n } \int _ { 0 } ^ { \infty }  & P\left(\sum _ { l = 1 } ^ { L_n }Z _ { il } > 0\right) ^ { \delta _ { L,i } }  P\left(\sum _ { l = 1 } ^ { L_n }Z _ { il } = 0, \sum _ { l = 1 } ^ { L_n }Y _ { il } > 0\right) ^ { \delta _ {I,i } }\\
		&\times P  \left(\sum _ { l = 1 } ^ { L_n }Z _ { il } = 0, \sum _ { l = 1 } ^ { L_n }Y _ { il } = 0\right) ^ { \delta _ {R,i } } f(\eta_i \mid r) {\rm d} \eta_i.
	\end{aligned}
	\label{obsLik-4}
\end{equation*}

Let  $p(A \mid \tau)$  represent  the probability mass function of a Poisson random variable $A$ with mean $\tau$.
Treating all latent variables as observable, the complete-data likelihood  is
\begin{equation}\tag{S2}
	\begin{aligned}
		\mathcal{L} _ { c } (\bbeta, \bgamma,\balpha) = \prod _ { i = 1 } ^ { n } \prod _ {l = 1 }^{L_n} & p(Z_{il} \mid \gamma _ { l } M _ { l } (t_{i1})\exp  \{ \bbeta ^ {\top}  \bX_i +   \phi_{\balpha}  ( \bW_i )\} \eta_i) \\
		&\times p(Y_{il} \mid \gamma _ { l } \{M _ { l } (t_{i2}) - M _ { l } (t_{i1})\}  \exp  \{ \bbeta ^ {\top}  \bX_i +   \phi_{\balpha}  ( \bW_i  )\} \eta_i)^{\delta _ {I , i } +\delta _ {R, i }}f(\eta_i \mid r),
	\end{aligned}
	\label{Lc}
\end{equation}
with constraints $\sum _ { l = 1 } ^ { L_n }Z _ { il } > 0$ if $\delta _ { L , i } = 1$,   $\sum _ { l = 1 } ^ { L_n }Z _ { il } = 0$ and $\sum _ { l = 1 } ^ { L_n }Y _ { il } > 0$ if $\delta _ {I, i } = 1$,  and  $\sum _ { l = 1 } ^ { L_n }Z _ { il } = 0$ and $\sum _ { l = 1 } ^ { L_n }Y _ { il } = 0$ if $ \delta _ {R , i }=1$.

In the E-step of the algorithm,  we calculate the  conditional expectation of $\log \{\mathcal{L} _ { c } (\bbeta, \bgamma, \balpha) \}$
with respect to all latent variables, given the observed data and  the $m$th parameter updates denoted by $(\bbeta^{(m)}, \bgamma^{(m)},\balpha^{(m)})$.
In particular, $(\bbeta^{(0)}, \bgamma^{(0)}, \balpha^{(0)})$ denotes the initial value of $(\bbeta, \bgamma, \balpha)$.
After  excluding some terms that are free of unknown parameters,
this step leads to the following objective function:
\begin{equation*}
	\begin{aligned}
		Q (\bbeta, \bgamma, \balpha;    \bbeta^{(m)}, & \bgamma^{(m)},\balpha^{(m)}) \\
		= \sum _ { i = 1 } ^ { n } \sum _ { l = 1 } ^ { L_n }& \Big( \{\operatorname { log } \gamma _ { l } + \bbeta ^ {\top} \bX _ { i } +   \phi_{\balpha} ( \bW_i )\} \{ E ( Z _ { i l } ) +( \delta _ { I , i } + \delta _ { R , i } )E ( Y _ { i l }  ) \}  \\
		& - \gamma _ { l } \exp  \{ \bbeta ^ {\top}  \bX_i + \phi_{\balpha}  ( \bW_i )\} E ( \eta _ { i } ) \{ ( \delta _ { L , i }+ \delta _ { I , i } ) M _ { l } ( R _ { i } ) + \delta _ { R , i } M _ { l } ( L _ { i } ) \}\Big).
		\label{Q}
	\end{aligned}	
\end{equation*}

For notational simplicity, we will ignore some arguments, including the current parameter update and observed data, in the below conditional expectations.
Let $\Lambda _ {\bgamma}^{(m)} (t) = \sum _ { l = 1 } ^ {L_n} \gamma _ { l }^{(m)} M _ { l } (t)$ and
$\mathcal{U}_i^{(m)}(t) = \Lambda _ {\bgamma}^{(m)} (t) $    $ \exp  \{\bX_i ^ {\top}  \bbeta^{(m)} + \phi_{\balpha}^{(m)} (\bW_i)\}$,
where   $m$ is a non-negative integer.
At the $m$th iteration of the proposed algorithm,
by utilizing the law of iterated expectations and  the facts that $Z_i =  \sum_{l=1}^{L_n}  Z_{il}$ and $Y_i =  \sum_{l=1}^{L_n}  Y_{il}$,
we have
$$
E(Z_{il}) = \frac{ \gamma _ { l }^{(m)} M _ { l } (R_i)}{\Lambda _ {\bgamma}^{(m)} (R_i)  } E(Z_i),
$$
and
$$
E(Y_{il}) = \frac{ \gamma _ { l }^{(m)} M _ { l } (R_i)-\gamma _ { l }^{(m)} M _ { l } (L_i)}{\Lambda _ {\bgamma}^{(m)} (R_i)  - \Lambda _ {\bgamma}^{(m)} (L_i) } E(Y_i).
$$

In addition, given $ \eta_{ i } $,  the conditional expectations  of $ Z_{ i } $ and  $ Y_{ i } $  are
$$
E(Z_{i} \mid  \eta _ { i }) = \frac{\mathcal{U}_i^{(m)}(R_i) \eta _ { i }   \delta_{L,i}}{1 - \exp \{ - \mathcal{U}_i^{(m)}(R_i)  \eta _ { i } \}},
$$
and
$$
E(Y_{i} \mid  \eta _ { i }) = \frac{ \{\mathcal{U}_i^{(m)} ( R_i ) \eta _ { i }   - \mathcal{U}_i^{(m)} (L_i)  \eta _ { i } \}  \delta_{I,i}}{1 - \exp \{ \mathcal{U}_i^{(m)} (L_i)  \eta _ { i } -  \mathcal{U}_i^{(m)} ( R_i )  \eta _ { i }  \}},
$$
respectively.
Again, with the law of iterated expectations,   the conditional expectations of $ Z_{ i } $ and  $ Y_{ i } $ are given by
$$
E(Z_{i}) = E_{\eta _ { i }} \{E(Z_{i} \mid  \eta _ { i }) \} = \frac{\mathcal{U}_i^{(m)}(R_i)  \delta_{L,i}}{1 - \exp  \{ - G [\mathcal{U}_i^{(m)}(R_i)  ] \}},
$$
and
\begin{equation*}
	\begin{aligned}
		E(Y_{i}) =& E_{\eta _ { i }} \{E(Y_{i} \mid  \eta _ { i })  \} \\
		= &\int_0^{\infty} \frac{ \{\mathcal{U}_i^{(m)} ( R_i ) \eta _ { i }   - \mathcal{U}_i^{(m)} (L_i)  \eta _ { i }  \}\delta_{I,i}}{1 - \exp  \{ \mathcal{U}_i^{(m)} (L_i)  \eta _ { i } -  \mathcal{U}_i^{(m)} ( R_i )  \eta _ { i }   \}} \\
		&\times \frac{ \exp  \{ -  \mathcal{U}_i^{(m)} ( L_i ) \eta _ { i }\} -
			\exp  \{ -  \mathcal{U}_i^{(m)} ( R_i ) \eta _ { i }\} }
		{ \exp  \{ - G [\mathcal{U}_i^{(m)} ( L_i ) ] \} -
			\exp  \{ -   G [ \mathcal{U}_i^{(m)} ( R_i ) ] \} }     f(\eta_i \mid r) {\rm d} \eta_i,
	\end{aligned}
\end{equation*}
respectively.
Specifically, the numerical Gauss-Laguerre method can be used to compute the aforementioned integral that lacks a closed form.
By using the Bayesian theorem,  the conditional expectation of $ \eta_{ i } $  takes the form
\begin{equation*}
	\begin{aligned}
		E ( \eta _ { i } ) = &\frac { \delta _ { L , i } \{ 1 - ( 1 + r \mathcal{U}_i^{(m)} (R_i) ) ^ { -(\frac { 1+r } { r } )} \} } {1 - \exp  \{ - G [ \mathcal{U}_i^{(m)} (R_i) ] \}}
		+  \frac { \delta _ { I , i } \{( 1 + r \mathcal{U}_i^{(m)} (L_i) ) ^ { -(\frac { 1+r } { r } )} -( 1 + r \mathcal{U}_i^{(m)} (R_i) ) ^ { -(\frac { 1+r } { r } )}\}}{ \exp  \{ - G [ \mathcal{U}_i^{(m)} (L_i) ]\} -
			\exp  \{ -   G [ \mathcal{U}_i^{(m)} (R_i)  ]  \} }  \\
		&+ \frac { \delta _ { R , i }( 1 + r \mathcal{U}_i^{(m)} (L_i) )^{  -(\frac { 1+r } { r } ) } } { \exp   \{ - G [\mathcal{U}_i^{(m)} (L_i) ] \}} .
	\end{aligned}
\end{equation*}


In the M-step of the algorithm, we proceed with updating the unknown parameter vector $\balpha$ in the DNN.
Specifically, we employ the popular mini-batch  SGD algorithm to   minimize the loss function
$-Q (\bbeta^{(m)}, \bgamma^{(m)}, \balpha;  \bbeta^{(m)}, \bgamma^{(m)},\balpha^{(m)})$.
In the $(m+1)$th iteration,    define $\phi_{\balpha}^{(m+1)} (\bW_i)$ as  the $(m+1)$th update  of  $\phi_{\balpha}  ( \bW_i  )$ after obtaining $\balpha^{(m+1)}$ for $i=1, \ldots, n$.
Since we impose the constraint   $E \{\phi ( \bW)\}=0$  to get an identifiable model,
we redefine $\phi_{\balpha}^{(m+1)} (\bW_i)$ as  $\phi_{\balpha}^{(m+1)}  ( \bW_i  ) -  \frac{1}{n}\sum_{i=1}^{n}\phi_{\balpha}^{(m+1)}( \bW_i  )$.
Next,  for each $l = 1 , \ldots , L_n$,  by setting  $\partial Q (\bbeta, \bgamma, \balpha^{(m+1)};  \bbeta^{(m)}, \bgamma^{(m)}, \balpha^{(m)})  / \partial \gamma _ { l } = 0$,
we can derive a closed-form update for $\gamma _ { l }$ that depends on the unknown $\bbeta$:
\begin{equation*}
	\begin{aligned}
		\gamma _ { l }^{(m+1)}(\bbeta; \balpha^{(m+1)}) = \frac { \sum _ { i = 1 } ^ { n } E ( Z _ { i l } ) + ( \delta _ { I , i } + \delta _ { R , i } )E ( Y _ { i l } )} { \sum _ { i = 1 } ^ { n }\exp  \{\bbeta ^ {\top}  \bX_i +  \phi_{\balpha}^{(m+1)} ( \bW_i )\}E(\eta_i) \{ ( \delta _ { L , i }+ \delta _ { I , i } ) M _ { l } ( R _ { i } ) + \delta _ { R , i } M _ { l } ( L _ { i } ) \} }.
	\end{aligned} 		
	\label{gammal}
\end{equation*}
Notably, given a nonnegative initial value for $\bgamma$, the proposed EM algorithm can automatically produce a nonnegative update of  $\bgamma$ at each iteration, ensuring a monotone update of cumulative baseline hazard.
This is a desirable feature of the proposed EM algorithm since it essentially avoids using the constrained optimization.
By replacing each  $\gamma _ { l }$  with $\gamma _ { l }^{(m+1)}(\bbeta; \balpha^{(m+1)})$  in $Q (\bbeta, \bgamma, \balpha^{(m+1)};  \bbeta^{(m)}, \bgamma^{(m)}, \balpha^{(m)})$,
we obtain the objective function for $\bbeta$ as follows:
\begin{equation*}
	\begin{aligned}
		Q_{\rm new}(\bbeta)  = \sum _ { i = 1 } ^ { n }& \sum _ { l = 1 } ^ { L_n }\Bigg(\{  \bbeta ^ {\top} \bX _ { i } +    \phi_{\balpha}^{(m+1)} ( \bW_i  )\} \times \{ E ( Z _ { i l } ) +( \delta _ { I , i } + \delta _ { R , i } )E ( Y _ { i l }  ) \} \\
		&- \log  \left[ { \sum _ { j = 1 } ^ { n }\exp \{\bbeta ^ {\top}  \bX_j +  \phi_{\balpha}^{(m+1)}( \bW_j  )\} E(\eta_j)\{ ( \delta _ { L , j }+ \delta _ { I , j } ) M _ { l } ( R _ { j } ) + \delta _ { R , j } M _ { l } ( L _ { j } ) \} } \right]   \\
		& \times \{ E ( Z _ { i l } ) +( \delta _ { I , i } + \delta _ { R , i } )E ( Y _ { i l }  ) \} \Bigg).
	\end{aligned}	
\end{equation*}
Then, the update  of  $\bbeta$ can be readily obtained by applying the existing
\texttt{R} optimization function, such as  \texttt{nlm()},  with  one step optimization
to  $Q_{\rm new}(\bbeta)$.

In summary, 	the proposed   algorithm proceeds with the following steps:

\noindent	
\textbf{Step 1:} Initialize  $\bbeta^{(0)}$, $\bgamma^{(0)}$,   and $\balpha^{(0)}$. Set $m=0$.

\noindent		
\textbf{Step 2:} In the $(m+1)$th iteration, compute all the conditional expectations,  $\{E ( \eta _ { i } ), E (Z_ { il} )$, $E (Y_ { il} ); i = 1, \ldots, n,  l = 1 , \ldots , L_n\}$, based on
$\bbeta^{(m)}$, $\bgamma^{(m)}$,  $\balpha^{(m)}$,
and the observed data.

\noindent		
\textbf{Step 3:} Apply SGD to  the loss function $-Q (\bbeta^{(m)}, \bgamma^{(m)}, \balpha;  \bbeta^{(m)}, \bgamma^{(m)}, \balpha^{(m)})$,
and obtain $\balpha^{(m+1)}$ and $\phi_{\balpha}^{(m+1)} (\bW_i)$ subsequently for $i = 1, \ldots, n$.

\noindent		
\textbf{Step 4:} Obtain $\boldsymbol{\beta }^{( m+ 1) }$ by applying the one step optimization to $Q_{\rm new}(\bbeta) $.

\noindent		
\textbf{Step 5:} Calculate $\gamma_l^{(m+1)}=\gamma_l(\bbeta^{(m+1)}; \balpha^{(m+1)})$ for $l = 1 , \ldots , L_n$.
Increase $m$ by 1.

\noindent	
\textbf{Step 6:} Repeat Step 2 to Step 5 until convergence is achieved.

In training the neural network, one often needs to tune several  hyperparameters, including the number of hidden layers $q$,  the number of neurons in each hidden layer, the tuning parameter in the  penalty function, batch size, epoch size, dropout rate, and learning rate \citep{bengio2017deep}.
These hyperparameters can be tuned via a grid search, evaluating the log-likelihood value on a hold-out validation set during each training trial.
Our experiences suggest that only the number of hidden layers, the tuning parameter in the  penalty function  and learning rate significantly impact the proposed method's performance. Thus, to save the computational burden, other insensitive hyperparameters are kept as fixed values in our numerical studies.


\section*{Section S.2: Proofs of Theorems 1--3}

For notational simplicity, we  define
$ \psi( \bm{U}) = \log \Lambda(R)+\phi(\bW)$, $ \psi(\bV)  = \log \Lambda(L)+\phi(\bW)$, $ \psi_{0}( \bm{U} ) = \log \Lambda_{0}(R)+\phi_{0}(\bW) $ and $ \psi_{0}(\bV) = \log \Lambda_{0}(L)+\phi_{0}(\bW) $, where   $ \bm{U} = (R, \bW^{\top})^{\top}$ and $ \bm{V} = (L, \bW^{\top})^{\top}$.
Define $\varUpsilon  (\bx, \by; \btheta) = G(\exp \{ \bbeta^{\top}\bx + \psi(\by)  \})  $ and
$\varUpsilon _{0}(\bx, \by) =G( \exp \{ \bbeta_{0}^{\top} \bx + \psi_{0}(\by) \} ) $.  
The log-likelihood for a single subject is given by 
\begin{equation*}
	\begin{aligned}
		\ell(\btheta) = &\delta_{L}\log[1-\exp(-\varUpsilon (\bX,\bU; \btheta))] +  \delta_{I}\log[\exp(-\varUpsilon (\bX,\bV; \btheta))-\exp(-\varUpsilon (\bX,\bU; \btheta))] \\
		&-\delta_{R}\varUpsilon (\bX,\bV; \btheta).
	\end{aligned}
\end{equation*}

Let $ \mathbb{P}_{n} $ denote the  empirical measure from $n$ independent observations and $\mathbb{P}$ denote the true probability measure.
Specifically, 
for a measurable function $ f $ 
and    a random variable $X$ with the distribution $F$,
$\mathbb{P}f $ is  defined as $\int f(x) {\rm d} F(x) $,   and $ \mathbb{P}_{n}f = \frac{1}{n} \sum\limits_{i=1}^n f(X_{i}) $,
where $X_{1}, \ldots, X_{n}$ are   $n$ independent realizations of $X$.  
Thus,   $ \mathbb{G}_{n}   = \sqrt{n}( \mathbb{P}_{n} -  \mathbb{P})$ is the corresponding  empirical process. 
Let  $\mathbb{M}(\btheta)=\mathbb{P}\{\ell(\btheta)\}$ and $\mathbb{M}_n(\btheta)=\mathbb{P}_n\{\ell(\btheta)\}$. 
Define $\btheta_n = (\bbeta, \Lambda_{\bgamma}, \phi_{\balpha})$,
$\mathcal{M}_B =\mathcal{M}(\mathcal{T}_{n}, d_M, B) $
and $  \mathcal{F}_B =\mathcal{F}(s,q,\boldsymbol{h},B)$ for some $B>0$. 
In what follows, $M$ denotes a general positive constant whose value may vary from place to place.   
Before proving Theorem 1, we   need the following lemma. 
\\

\newtheorem{lem}{\bf Lemma}

\noindent	
{\bf Lemma 1.}
{\sl
	Under  conditions (C2)--(C4),    the class $ \{\ell(\btheta_n): \btheta_n \in \mathcal{D}  \times \mathcal{M}_B  \times \mathcal{F}_B \}$ is   Glivenko-Cantelli}.\\

\noindent	
{\bf  Proof.}	
For any $\varepsilon >0$, we first know that
there are $O(\varepsilon^{-p})$ number of brackets  covering the compact set $\mathcal{D}$, such that the difference of any two $\bbeta$
within the same bracket 	under  Euclidean
norm is at most $\varepsilon$.
Consequently, the $\varepsilon$-bracketing number for the class $ \{\bbeta^{\top}\bX: \bbeta \in  \mathcal{D}\}$  equipped by  Euclidean
norm is $O(\varepsilon^{-p})$ by  Condition (C2).
Similarly, there exist $O((B / \varepsilon)^{L_n})$
$ \varepsilon$-brackets to cover
$\{\bgamma \in \mathbb{R}^{L_n} : 0 \leq \gamma_{ l } \leq B, l = 1, \ldots, L_n\}$.
Then,  by Lemma 2.5 in \citep{van2000applications},
the $\varepsilon$-bracketing number of  $\mathcal{M}_B$ equipped by  Euclidean
norm  is  bounded by $O((B/\varepsilon)^{L_n}) $ under Conditions (C3) and (C4).

To calculate the entropy of	  $\mathcal{F}_B $, we 
apply  similar arguments as demonstrated in the proof of Lemma 6 of \cite{zhong2022deep}.
Consider two neural networks $\phi_{1}$ and $\phi_{2} \in $  $\mathcal{F}_B$, 
and let 	  $\balpha_{\phi_{1}} = (\bomega_{1,q}, \bv_{1,q}, \ldots, \bomega_{1,0}, \bv_{1,0})$ and $\balpha_{\phi_{2}} = (\bomega_{2,q}, \bv_{2,q}, \ldots, \bomega_{2,0}, \bv_{2,0})$ be two sets of parameters related to   $\phi_{1}$ and $\phi_{2}$, respectively. 	
We define the supremum norm of $(\balpha_{\phi_{1}}-\balpha_{\phi_{2}})$ as follows 
$$
\left\|\balpha_{\phi_{1}}-\balpha_{\phi_{2}}\right\|_{\infty}=\max _{j=0, \ldots,q} \left(\left\|\bomega_{1,j}-\bomega_{2,j}\right\|_{\infty} \vee \left\|\bv_{1,j}-\bv_{2,j}\right\|_{\infty}\right).
$$

Next, define the neural networks $\phi_{1,2}^j$ with parameters given by 
$$
\balpha_{\phi_{1,2}^j}=\left(\bomega_{1,q}, \bv_{1,q}, \ldots, \bomega_{1, j+1}, \bv_{1, j+1}, \bomega_{2,j}, \bv_{2,j}, \ldots, \bomega_{2,0}, \bv_{2,0}\right), \text { for } j=0, \ldots,  q-1.
$$
According to the neural network's definition, we have
$$
\begin{aligned}
	\left\|\phi_{1,2}^j-\phi_{1,2}^{j-1}\right\|_{\infty}  \leq\left(\prod_{k=j+1}^q h_k\right)\left\{\left\|\left(\bomega_{2,j}-\bomega_{1,j}\right) \phi_{j}^*\right\|_{\infty}+\left\|\bv_{1,j}-\bv_{2,j}\right\|_{\infty}\right\},
\end{aligned}
$$
where $\phi_{j}^*$ is a neural network with parameters $\balpha_{\phi_{j}^*}=\left({\rm \bI}_j, \bv_{2,j}, \bomega_{2,j-1}, \bv_{2,j-1}, \ldots, \bomega_{2,0}, \bv_{2,0}\right)$ and ${\rm \bI}_j$ represents an identity matrix. 
Since   $\left\|\phi_{j}^*\right\|_{\infty}$ is bounded by $\prod_{k=0}^{j-1}\left(h_k+1\right)$,   we can conclude that
$$
\begin{aligned}
	\left\|\phi_{1,2}^j-\phi_{1,2}^{j-1}\right\|_{\infty}   \leq\left\{\prod_{k=0}^q\left(h_k+1\right)\right\}\left\|\balpha_{\phi_{1}}-\balpha_{\phi_{2}}\right\|_{\infty}.
\end{aligned}
$$
Furthermore, we have
$$
\left\|\phi_{1}-\phi_{2}\right\|_{\infty} \leq \sum_{j=1}^q\left\|\phi_{1,2}^j-\phi_{1,2}^{j-1}\right\|_{\infty} \leq q\left\{\prod_{k=0}^q\left(h_k+1\right)\right\}\left\|\balpha_{\phi_{1}}-\balpha_{\phi_{2}}\right\|_{\infty} .
$$

Given the fact that a neural network has at most $K_{h}=\sum_{j=0}^q\left(h_j+1\right) h_{j+1}$ parameters, there are $\binom{K_{h}}{\tilde{s}}$  possible ways to select $\tilde{s}  \, (\leq s)$ non-zero parameters. 
Define $K=q\prod_{k=0}^q (h_k+1) \sum_{k=0}^q h_k h_{k+1}$.
Therefore, for any $\varepsilon>0$, the covering number $N \left(\varepsilon, \mathcal{F}_B,L_{1}(\mathbb{P})\right) $ satisfies
$$
N \left(\varepsilon, \mathcal{F}_B,L_{1}(\mathbb{P})\right)  \lesssim \sum_{\tilde{s}=1}^s\binom{K_{h}}{\tilde{s}}\left\{\frac{q \prod_{k=0}^q\left(h_k+1\right)}{\varepsilon}\right\}^{\tilde{s}} \lesssim\left(\frac{K}{\varepsilon}\right)^{s+1} .
$$
By   Theorem 2.7.11 in \cite{van1996weak}, we can obtain that the $\varepsilon$-bracketing number of $\mathcal{F}_{B} $ equipped by $L_2$-norm is  bounded by $O((K/\varepsilon)^{s})$.

For any $\btheta_1$ and $\btheta_2 \in \mathcal{D}  \times \mathcal{M}_B  \times \mathcal{F}_B$,
by using Taylor series expansion,  we get
$$
\Vert \ell(\btheta_1) -  \ell(\btheta_2) \Vert_{L_2}^{2} \lesssim \lVert \bbeta_1 - \bbeta_{2} \lVert^{2} + \lVert \Lambda _1- \Lambda_{2} \lVert_{L_{2}}^{2} + \lVert \phi_1 - \phi_{2} \lVert_{L_{2}}^{2}.
$$
Then the $\varepsilon$-bracketing number of the class $ \{\ell(\btheta_n): \btheta_n \in \mathcal{D}  \times \mathcal{M}_B  \times \mathcal{F}_B \}$, equipped by $d(\btheta_1, \btheta_2)$  defined in Section 3 of the main paper,  is bounded by $M(1/\varepsilon)^p (B/\varepsilon)^{L_n} (K/\varepsilon)^s$.
By Theorem 2.5.6 in \cite{van1996weak}, we  can conclude that   the class $ \{\ell(\btheta_n): \btheta_n \in \mathcal{D}  \times \mathcal{M}_B  \times \mathcal{F}_B \}$ is Donsker and thus    Glivenko-Cantelli.
\\

\noindent	
{\bf Lemma 2.}
{\sl
	Under  conditions (C1)--(C3), we have
	$$	\mathbb{M}(\btheta_0) -\mathbb{M}(\btheta) \asymp  d^{2}(\btheta, \btheta_{0}),$$
	for all $\btheta$ in the set  $ \{\btheta \in \mathcal{D}  \times \mathcal{M}_B  \times \mathcal{F}_B : d(\btheta, \btheta_0) < \delta\}$, where $\delta > 0$ is sufficiently small.}\\

\noindent	
{\bf  Proof.}	We first have $$
\begin{aligned}
	\mathbb{M}&(\btheta_0) -\mathbb{M}(\btheta)  = E\left([1-\exp \{-\varUpsilon(\bX, \bU; \btheta)\}] m\left[\frac{1-\exp \{-\varUpsilon_0(\bX, \bU)\}}{1-\exp \{-\varUpsilon(\bX, \bU; \btheta)\}}\right]\right.\\
	&+[\exp \{-\varUpsilon(\bX, \bV; \btheta)\}-\exp \{-\varUpsilon(\bX, \bU; \btheta)\}] m\left[\frac{\exp \{-\varUpsilon_0(\bX, \bV)\}-\exp \{-\varUpsilon_0(\bX, \bU)\}}{\exp \{-\varUpsilon(\bX, \bV; \btheta)\}-\exp \{-\varUpsilon(\bX, \bU; \btheta)\}}\right] \\
	&\left.+\exp \{-\varUpsilon(\bX, \bV; \btheta)\} m\left[\frac{\exp \{-\varUpsilon_0(\bX, \bV)\}}{\exp \{-\varUpsilon(\bX, \bV; \btheta)\}}\right]\right),
\end{aligned}
$$
where $m(x)=x \log x - x+1$  and  $(x - 1)^2 / 4 \leq m(x)  \leq (x -1)^2$ if $x \rightarrow$  1. 
Define  $h(z) = 1- \exp\{-\varUpsilon(\bX, z; \btheta)\}, h_0(z) = 1- \exp\{-\varUpsilon_0(\bX, z)\}$, and   $ \mathscr{G} (\bU, \bV; \btheta)= (\varUpsilon(\bX, \bU; \btheta) - \varUpsilon_{0}(\bX, \bU))^{2} + (\varUpsilon(\bX, \bV; \btheta) - \varUpsilon_{0}(\bX, \bV))^{2} $. 
By Conditions (C1)--(C3),  we utilize the Taylor series expansion and  derive
$$
\begin{aligned}
	\mathbb{M}&(\btheta_0)-\mathbb{M}(\btheta) =\\ 
	&E\left\{ h(\bU) m \left(\frac{h_0(\bU)}{h(\bU)}\right) +(h(\bU)-h(\bV))m \left(\frac{h_0(\bU)-h_0(\bV)}{h(\bU)-h(\bV)}\right) + (1-h(\bV)) m\left(\frac{1-h_0(\bV)}{1-h(\bV)}\right)\right\}\\			
	\gtrsim &  E\left\{\frac{1}{h(\bU)}[h(\bU)-h_0(\bU)]^2+\frac{1}{1-h(\bV)}[h(\bV)-h_0(\bV)]^2\right\} \\
	\gtrsim &  E\left\{\frac{1}{\varUpsilon(\bX, \bU; \btheta)}[\varUpsilon(\bX, \bU; \btheta)-\varUpsilon_0(\bX, \bU)]^2+\frac{1}{1-\varUpsilon(\bX, \bV; \btheta)}[\varUpsilon(\bX, \bV; \btheta)-\varUpsilon_0(\bX, \bV)]^2\right\} \\
	\gtrsim &  E \{\mathscr{G} (\bU, \bV; \btheta)\}.
\end{aligned}
$$
Under Conditions (C1) and (C2),  by Lemma 25.86 of \cite{van2000asymptotic} and the mean value theorem,
we have 
$$
\begin{aligned}
	E \{\varUpsilon(\bX, \bU; \btheta) - \varUpsilon_{0}(\bX, \bU)\}^{2}  
	& \gtrsim E\{ \lVert \bbeta - \bbeta_{0} \lVert^{2} + \lVert \Lambda_{\gamma}(R) - \Lambda_{0}(R) \lVert_{L_{2}}^{2} + \lVert \phi_{\balpha}(\bW) - \phi_{0}(\bW) \lVert_{L_{2}}^{2}\} \\
	& =d^{2}(\btheta, \btheta_{0}).
\end{aligned}
$$
Similarly, $E \{\varUpsilon(\bX, \bV; \btheta) - \varUpsilon_{0}(\bX, \bV)\}^{2}  \gtrsim d^{2}(\btheta, \btheta_{0})$. Then we have $E \{\mathscr{G} (\bU, \bV; \btheta)\} \gtrsim d^{2}(\btheta, \btheta_{0})$.
By  the inequality $ m(x) \leq (x-1)^2 $,
we  apply  similar  arguments as above to show that $\mathbb{M}(\btheta_0)-\mathbb{M}(\btheta) \lesssim E \{\mathscr{G} (\bU, \bV; \btheta)\} \lesssim d^{2}(\btheta, \btheta_{0})$.
Thus, we have
$$
\mathbb{M}(\btheta_0) -\mathbb{M}(\btheta) \asymp  d^{2}(\btheta, \btheta_{0}).
\label{lem2}
$$
\\ 

\noindent
{\bf Proof of Theorem 1.}
To establish the consistency of $\hat{\btheta}$,  we proceed with   the following three steps.
First, 
since the class of functions $ \{\ell(\btheta_n): \btheta_n \in \mathcal{D}  \times \mathcal{M}_B  \times \mathcal{F}_B \}$ is shown to be  Glivenko-Cantelli in Lemma 1, we have
\begin{equation}\tag{S3}
	\sup\limits_{\btheta_n \in \mathcal{D}  \times \mathcal{M}_B  \times \mathcal{F}_B } | \mathbb{M}_{n}(\btheta_n) - \mathbb{M}(\btheta_n )| = o_{p}(1) .
	\label{Con1}
\end{equation}
Second, we show that $\sup\limits_{\btheta_n : d(\btheta_n,\btheta_0)\geq \delta} \mathbb{M}(\btheta_n ) < \mathbb{M}(\btheta_0 ) $ for every $\delta>0$, which   is satisfied if we can prove the model  identifiability.
In particular, for any $\btheta_1$ and $\btheta_2 \in \mathcal{D}  \times \mathcal{M}_B  \times \mathcal{F}_B$, if $\ell(\btheta_1) = \ell(\btheta_2)$ with probability one, we need to show that $\btheta_1 = \btheta_2$.
To this end, choosing $\delta_L = 1$ in $\ell(\btheta_1)$ and $\ell(\btheta_2)$,
we   obtain $\exp\{\bbeta_{1}^{\top} \bX + \phi_{1}(\bW)\}  \Lambda_{1}(L)=\exp\{\bbeta_{2}^{\top} \bX + \phi_{2}(\bW)\}  \Lambda_{2}(L)$. 
Thus, for any $t \in [a, b]$ and $\bW \in \mathbb{R}^{d}$,
$$\exp\{\bbeta_{1}^{\top} \bX + \phi_{1}(\bW)\}  \Lambda_{1}(t)=\exp\{\bbeta_{2}^{\top} \bX + \phi_{2}(\bW)\}  \Lambda_{2}(t).
$$   
Taking the logarithm of both sides of the above equation leads to
\begin{equation}\tag{S4} 
	\bbeta_{1}^{\top} \bX + \phi_{1}(\bW)   + \log\{\Lambda_{1}(t)\}= \bbeta_{2}^{\top} \bX + \phi_{2}(\bW)   + \log\{\Lambda_{2}(t)\}.
	\label{S3} 
\end{equation} 
Further,
taking expectation of both sides of the above equation with respect to $\bW$ and using the fact    $E\{\phi_{1}(\bW)\} =  E\{\phi_{2}(\bW)\} =0$,
we have
$$ \bbeta_{1}^{\top} \bX   + \log\{\Lambda_{1}(t)\}= \bbeta_{2}^{\top} \bX   + \log\{\Lambda_{2}(t)\}.
$$   
By Condition (C8), we have  $\bbeta_{1} = \bbeta_{2}$ and $\Lambda_{1}(t)=\Lambda_{2}(t)$ for $t \in [a, b]$.
By this conclusion and equation \eqref{S3}, we have $\phi_{1}(\bW) =  \phi_{2}(\bW)$ for any $\bW \in \mathbb{R}^{d}$.
Thus, the proposed model is identifiable, and we have
\begin{equation}\tag{S5} 
	\sup\limits_{\btheta_n : d(\btheta_n,\btheta_0)\geq \delta} \mathbb{M}(\btheta_n ) < \mathbb{M}(\btheta_0 ).
	\label{Con2} 
\end{equation}

Third, we show that $\mathbb{M}_n(\hat{\btheta}) \geq \mathbb{M}_n(\btheta_0)  - o_p(1). $
Define $\mathcal{M}_{B/2} = \{ \sum_{l=1}^{L_n}  \gamma_{ l }M_l(t) :  0 \leq \gamma_{ l } \leq B/2, \text{ for }  l=1, \ldots, L_n, t \in [a,b] \}$,   $  \mathcal{F}_{B/2} =\mathcal{F}(s,q,\boldsymbol{h},B/2)$,
$ \Lambda_{0,n}=\underset{\Lambda_{\bgamma} \in  \mathcal{M}_{B/2} } {\operatorname*{\arg\min}}\lVert \Lambda_{\bgamma}- \Lambda_0\rVert_{L_2} $ and $ \phi_{0,n}=\underset{\phi_{\balpha} \in  \mathcal{F}_{B/2} } {\operatorname*{\arg\min}}\lVert \phi_{\balpha}- \phi_0\rVert_{L_2} $.
By Condition (C4), according to Corollary 6.21 of \cite{schumaker2007spline} and Lemma A1 of \cite{lu2007estimation}, 
we have $\lVert \Lambda_{0,n}- \Lambda_0\rVert_{L_2} = O(n^{-\xi\nu})$,  where $ 1/(2\xi + 1) < \nu < 1/(2\xi) $.
Furthermore,  according to Theorem 1 of \cite{schmidt2020nonparametric}, we have $\lVert \phi_{0,n}- \phi_0\rVert_{L_2} = O(\alpha_n \log^2n)$.

Define $ \btheta_{0,n}= (\bbeta_{0},\Lambda_{0,n},\phi_{0,n})$ and $ \btheta_{0,n}^{\prime}= (\bbeta_{0},\Lambda_{0},\phi_{0,n})$. 
By (\ref{Con1}), Lemma 2 and the law of large numbers,  we have
$$
\begin{aligned}
	|\mathbb{M}_n(\btheta_{0,n})-\mathbb{M}_n(\btheta_{0})|&\leq |\mathbb{M}_n(\btheta_{0,n})-\mathbb{M}(\btheta_{0,n})|+|\mathbb{M}(\btheta_{0,n})-\mathbb{M}(\btheta_{0,n}^{\prime})|\\
	&+|\mathbb{M}(\btheta_{0,n}^{\prime})-\mathbb{M}(\btheta_{0})|+|\mathbb{M}(\btheta_{0})-\mathbb{M}_n(\btheta_{0})|\\
	&=o_p(1).
\end{aligned}
$$
Then, we can obtain 
$$
\mathbb{M}_n( 	\hat{\boldsymbol{\theta}}) - \mathbb{M}_n(\btheta_{0})  \geq \mathbb{M}_n(\btheta_{0,n}) - \mathbb{M}_n(\btheta_{0})  \geq - o_p(1),
$$
which gives 
\begin{equation}\tag{S6}
	\mathbb{M}_n( 	\hat{\boldsymbol{\theta}}) \geq  \mathbb{M}_n(\btheta_{0}) - o_p(1).
	\label{Con3}
\end{equation} 
Based on the uniform convergence of $\mathbb{M}_n$ to $\mathbb{M}$  as obtained from   \eqref{Con1}, we have 
$\mathbb{M}_n\left(\btheta_0\right) \xrightarrow{p} \mathbb{M}\left(\btheta_0\right)$ and $\mathbb{M}_n( 	\hat{\boldsymbol{\theta}}) \geq  \mathbb{M}(\btheta_{0}) - o_p(1)$. The latter implies  
$$
\begin{aligned}
	\mathbb{M}\left(\btheta_0\right)-\mathbb{M}(\hat{\btheta}) & \leq \mathbb{M}_n(\hat{\btheta})-\mathbb{M}(\hat{\btheta})+o_p(1) \\
	& \leq \sup\limits_{\btheta_n \in \mathcal{D}  \times  \mathcal{M}_B  \times \mathcal{F}_B} | \mathbb{M}_{n}(\btheta_n) - \mathbb{M}(\btheta_n )| +o_p(1) \xrightarrow{p} 0
\end{aligned}
$$
According to the  inequality \eqref{Con2},  
for every $\delta>0$, there exists a positive number $c_{\theta}>0$ such that $\mathbb{M}(\btheta)<\mathbb{M}(\btheta_0)-c_{\theta}$ for any $\btheta$ with $d\left(\btheta, \btheta_0\right) \geq \delta$.
Clearly,  the event $\{d(\hat{\btheta}, \btheta_0) \geq \delta\}$ is contained in the event $\{\mathbb{M}(\hat{\btheta})<\mathbb{M}(\btheta_0)-c_{\theta}\}$, and the probability of the event $\{\mathbb{M}(\hat{\btheta})<\mathbb{M}(\btheta_0)-c_{\theta}\}$ converges to 0.
Hence, we have    $ d(\hat{\btheta}, \btheta_{0})  =o_p(1) $.
\\

\noindent	
{\bf Proof of Theorem 2.}
For $\delta >0$, define $ \bm{\Theta}_{\delta} = \{\btheta_n \in  \mathcal{D}  \times \mathcal{M}_B  \times \mathcal{F}_B, \delta/2 \leq d(\btheta_n, \btheta_{0}) \leq \delta\}$. 
From Lemma 2,  we have $\mathbb{M}(\btheta_0) -\mathbb{M}(\btheta) \gtrsim d^2(\btheta, \btheta_{0}) \gtrsim \delta^2$,   leading to
\begin{equation*}
	\sup_{\btheta_n\in  \bm{\Theta}_{\delta}} [\mathbb{M}(\btheta_n)-\mathbb{M}(\btheta_0)] \lesssim -\delta^2.
	\label{cond1}
\end{equation*}
Recall that   $ \btheta_{0,n}= (\bbeta_{0},\Lambda_{0,n},\phi_{0,n})$,
where  $ \Lambda_{0,n}=\underset{\Lambda_{\gamma} \in  \mathcal{M}_{B/2} } {\operatorname*{\arg\min}}\lVert \Lambda_{\gamma}- \Lambda_0\rVert_{L_2} $ and  $ \phi_{0,n}=\underset{\phi_{\balpha} \in  \mathcal{F}_{B/2} } {\operatorname*{\arg\min}}\lVert \phi_{\balpha}- \phi_0\rVert_{L_2} $.
Let $ \mathcal{L}_{\delta} = \{\ell(\btheta_n) - \ell(\btheta_{0,n}): \btheta_n \in \mathcal{D}  \times \mathcal{M}_B  \times \mathcal{F}_B, d(\btheta_n, \btheta_{0,n}) \leq \delta\}$.
We have
$$ \lVert  \mathbb{G}_n \rVert_{\mathcal{L}_{\delta}} =  \mathop{\sup}\limits_{\substack{\btheta_n \in \mathcal{D}  \times \mathcal{M}_B \times \mathcal{F}_B, \\ d(\btheta_n, \btheta_{0,n})\leq \delta }}  |\mathbb{G}_n \ell(\btheta_n)  - \mathbb{G}_n \ell(\btheta_{0,n})| .
$$

By Condition (C1) and Theorem 9.23 of \cite{kosorok2008introduction}, for any $\delta >0 $ and $0<\varepsilon<\delta$, the $\varepsilon$-bracketing number of $\{\bbeta \in \mathcal{D} : \lVert \bbeta - \bbeta_{0} \lVert \leq \delta \}$ with radius $\varepsilon$  and Euclidean  norm  is bounded by $M(\delta/\varepsilon)^p$.
By Condition(C4), following the proof of  Lemma 1  and the calculations of \cite{shen1994convergence} (pp. 597),
we  conclude that the logarithms of the $\varepsilon$-bracketing numbers of  $ \mathcal{M}_{\delta} = \{\Lambda_{\bgamma} \in  \mathcal{M}_B:  \Vert \Lambda_{\bgamma} - \Lambda_{0,n} \Vert_{L_2} \leq \delta\}$ and $ \mathcal{F}_{ \delta} = \{\phi_{\balpha} \in  \mathcal{F}_B: \Vert \phi_{\balpha} - \phi_{0,n} \Vert_{L_2} \leq \delta\}$ equipped with $ L_{2}(\mathbb{P}) $  are bounded by $ M L_n \log(\delta/\varepsilon) $ and $Ms \log (K/\varepsilon)$, respectively. 
Therefore, it follows that if $p  \leq s$ and $\delta \leq K$,
$$
\log N_{[]}(\varepsilon, \mathcal{L}_{\delta},  L_{2}(\mathbb{P}) ) \lesssim p \log(\delta/\varepsilon)+L_n \log(\delta/\varepsilon) +s\log(K/\varepsilon) \lesssim
(s+L_n)\log(K/\varepsilon).
$$

Define	the entropy integral   $ J_{[ ] }(\delta, \mathcal{F}, L_{2}(\mathbb{P})) $   as $ \int_{0}^{\delta}\sqrt{1+\log N_{[ ] }(\varepsilon, \mathcal{F}, L_{2}(\mathbb{P}) )}d\varepsilon$. 
We obtain  
$$
\begin{aligned}
	J_{[]}(\delta,  \mathcal{L}_{\delta}, L_2(\mathbb{P})) & \lesssim
	\int_{0}^{\delta}\sqrt{1+(s+L_n)\log(K/\varepsilon)}d\varepsilon\\
	&=\frac{2K}{(s+L_n)}e^{1/(s+L_n)}\int_{\sqrt{1+(s+L_n)\log(K/\delta)}}^{\infty}u^2\exp^{-u^2/(s+L_n)} du\\
	&\asymp\delta\sqrt{(s+L_n) \log K/\delta}.\nonumber
\end{aligned}
$$
The second equality holds by setting  $u=\sqrt{1+(s+L_n)\log(K/\varepsilon)}$.
The last inequality holds due to the fact that $u>\sqrt{(s+L_n)\log(K/\varepsilon)}$.
Referring to the  Lemma 3.4.3 in  \cite{van1996weak}, we have
$$
\begin{aligned}
	\mathbb{E}^*\|\mathbb{G}_n\|_{\mathcal{L}_\delta}&\lesssim J_{[]}(\delta,\mathcal{L}_\delta,L_2(\mathbb{P}))\left\{1+\frac{J_{[]}(\delta,\mathcal{L}_\delta,L_2(\mathbb{P}))}{\delta^2\sqrt{n}}\right\}\\&\lesssim\delta\sqrt{(s+L_n)\log(K/\delta)}+\frac {(s+L_n)}{\sqrt{n}}\log(K/\delta).
\end{aligned}$$

To derive the convergence rate of the proposed estimator, we let
the pivotal function $\phi_{n}(\delta)$ defined in Theorem 3.2.5 of \cite{van1996weak}  take  the form $\phi_{n}(\delta)=\delta\sqrt{(s+L_n)\log(K/\delta)}+\frac{(s+L_n)}{\sqrt{n}}\log(K/\delta).$ 
Based on the preceding results  $\|\phi_{0,n}-\phi_0\|_{L_2}=O(\alpha_n\log^2n)$ and $\|\Lambda_{0,n}-\Lambda_0\|_{L_2}=O(n^{-\xi\nu})$, we obtain
\begin{equation*}
	\mathbb{E}^*\sup_{\btheta_n \in  \bm{\Theta}_{\delta}}  \sqrt{n}|(\mathbb{M}_n-\mathbb{M})(\btheta_n)-(\mathbb{M}_n-\mathbb{M})(\btheta_0)| \lesssim \phi_n(\delta).
\end{equation*}
We	 set  $r_n $ defined in Theorem 3.2.5 of \cite{van1996weak} to   $r_n =( \alpha_n\log^2n+n^{-\xi \nu})^{-1}$.
By Condition  (C2), Condition (C6) and $p_n=O(n^\nu)$,
we have
\begin{equation*}
	r_n^{2} \phi_{n}(1/r_n)= \frac{\frac{1}{r_n}\sqrt{(s+L_n)\log(Kr_n)}+\frac{(s+L_n)}{\sqrt{n}}\log(Kr_n)}{ (\alpha_n\log^2n+n^{-\xi\nu})^2} \leq \sqrt{n} .
\end{equation*}
Therefore, we can choose $r_n = ( \alpha_n\log^2n+n^{-\xi\nu})^{-1}$ such that 
\begin{equation*}
	\begin{aligned}
		| \mathbb{M}_n(\btheta_{0,n})-\mathbb{M}_n(\btheta_0) | &\lesssim |(\mathbb{M}_n-\mathbb{M})(\btheta_{0,n})-(\mathbb{M}_n-\mathbb{M})(\btheta_0)| +|\mathbb{M}(\btheta_{0,n})-\mathbb{M}(\btheta_0)| \\
		&\lesssim O_p( \phi_{n}(1/r_n)/\sqrt{n})+\|\phi_{0,n}-\phi_0\|_{L_2}^2 +\|\Lambda_{0,n}-\Lambda_0\|_{L_2}^2\\
		&=O_p(r_n^{-2}).
	\end{aligned}
\end{equation*}
By the definition of $\hat{\boldsymbol{\theta}}$,  we derive the inequalities 
$$
\mathbb{M}_n( 	\hat{\boldsymbol{\theta}}) - \mathbb{M}_n(\btheta_{0}) \geq \mathbb{M}_n(\btheta_{0,n}) -\mathbb{M}_n(\btheta_{0}) \geq  - O_p(r_n^{-2 })  ,
$$
According to Theorem 3.2.5 from \cite{van1996weak}, it follows that
$r_n d(\hat{\boldsymbol{\theta}},\boldsymbol{\theta}_0)=O_p(1).$
In other words,  $d(\hat{\boldsymbol{\theta}},\boldsymbol{\theta}_0) = O_p(\alpha_n\log^2n+n^{-\xi\nu})$.
\\

\noindent	
{\bf Proof of Theorem 3.} Define 
$$
\begin{aligned}
Q_1(\bX,\bU,\bV; \btheta) = & \delta_{L}\left(  \frac{\exp \{-\varUpsilon(\bX, \bU; \btheta)\} G^{\prime}[\exp\{\bbeta^{\top}\bX + \psi( \bm{U})\}]   \exp\{\bbeta^{\top}\bX + \psi( \bm{U})\}}{1-\exp \{-\varUpsilon(\bX, \bU; \btheta)\}} \right)\\
&+\delta_{I} \left( \frac{\exp \{-\varUpsilon(\bX, \bU; \btheta)\}G^{\prime}[\exp\{\bbeta^{\top}\bX + \psi( \bm{U})\}]  \exp\{\bbeta^{\top}\bX + \psi( \bm{U})\})}{\exp \{-\varUpsilon(\bX, \bV; \btheta)\} - \exp \{-\varUpsilon(\bX, \bU; \btheta)\}}  \right),
\end{aligned}
$$
$$
\begin{aligned}
Q_2(\bX,\bU,\bV; \btheta) =& -\delta_{I} \left( \frac{\exp \{-\varUpsilon(\bX, \bV; \btheta)\}G^{\prime}[\exp\{\bbeta^{\top}\bX + \psi( \bm{V})\}]  \exp\{\bbeta^{\top}\bX + \psi( \bm{V})\} }{\exp \{-\varUpsilon(\bX, \bV; \btheta)\} - \exp \{-\varUpsilon(\bX, \bU; \btheta)\}}  \right) \\
&-\delta_{R} \left(  G^{\prime}[\exp\{\bbeta^{\top}\bX + \psi( \bm{V})\}] \exp\{\bbeta^{\top}\bX + \psi( \bm{V})\} \right).
\end{aligned}
$$
The score function for $\bbeta$ is 
$$
\begin{gathered}
\dot{\ell}_{\boldsymbol{\beta}}(\btheta_0)=\left.\frac{\partial \ell(\boldsymbol{\beta}, \Lambda_{0}, \phi_{0} ))}{\partial \boldsymbol{\beta}}\right|_{\bbeta=\bbeta_0}=
\bX (Q_1(\bX,\bU,\bV; \btheta_0)+Q_2(\bX,\bU,\bV; \btheta_0)).
\end{gathered}
$$
Consider any parametric submodel of $\Lambda$  denoted as $ \Lambda_{\kappa,q}(t)  = \Lambda(t)  + \kappa q (t) $,  where $q  \in L_2([a,b])$. The score function along this submodel is 
$$
\dot{\ell}_{\Lambda}(\btheta_0)[q]=\left.\frac{\partial \ell(\boldsymbol{\beta}_0, \Lambda_{\kappa,q}, \phi_{0} )}{\partial \kappa}\right|_{\kappa=0} = \frac{q(R)}{\Lambda(R)} Q_1(\bX,\bU,\bV ; \btheta_0)+\frac{q(L)}{\Lambda(L)} Q_2(\bX,\bU,\bV; \btheta_0).
$$
To derive the score operator for $\phi$, we 
consider any parametric submodel of $\phi$  given by
$ \phi_{\zeta,z}(\bw) = \phi(\bw) +\zeta z (\bw)$,  where $z  \in L_2([e,f]^d)$. The score function along this submodel is 
$$
\dot{\ell}_{\phi}(\btheta_0)[z]=\left.\frac{\partial \ell(\boldsymbol{\beta}_0, \Lambda_{0}, \phi_{\zeta,z})}{\partial \zeta}\right|_{\zeta=0}=z(\bw) (Q_1(\bX,\bU,\bV ; \btheta_0)+Q_2(\bX,\bU,\bV ; \btheta_0)).
$$
Let $ \Psi_{\Lambda_0} $ be the collection of all subfamilies $\{\Lambda_{_{\kappa,q}}: \kappa\in(-1,1) \} \subset \{\Lambda:\Lambda \in \bar{\Psi}_{\Lambda}\}$ such that $\lim_{\kappa\to0} \| \kappa^{-1}(\Lambda_{\kappa,q}-\Lambda_0)-q \| _{L_2}=0$, where $q \in L_2([a,b])$, and let
$$
\begin{aligned}
\Omega_{\Lambda_0}=&\{q \in L_2([a,b]) : \lim_{\kappa \to 0} \| \kappa^{-1}(\Lambda_{\kappa,q}-\Lambda_0)-q \| _{L_2}=0 \text{ for some subfamily} \\
&\{\Lambda_{_{\kappa,q}}: \kappa\in(-1,1) \} \in \Psi_{\Lambda_0}\}.
\end{aligned}
$$

Similarly, let $\Psi_{\phi_0}$ denote the collection of all subfamilies $\{\phi_{\zeta,z} \in L_2([e,f]^d : \zeta \in(-1,1)) \} \subset 	\bar{\Psi}_{\phi}$  such that $\lim_{\zeta \to0} \| \zeta^{-1}(\phi_{\zeta,z}-\phi_0)-z \| _{L_2}=0$, where $z \in L_2([e,f]^d)$, and let
$$
\begin{aligned}
\Omega_{\phi_0}=&\{z \in L_2([e,f]^d) : \lim_{\zeta \to 0} \| \zeta^{-1}(\phi_{\zeta,z}-\phi_0)-z \| _{L_2}=0 \text{ for some subfamily} \\
&\{\phi_{_{\zeta,z}}: \zeta \in(-1,1) \} \in \Psi_{\phi_0}, \text{ and } E\{z(\bW)\}=0\}.
\end{aligned}
$$
Let $\overline{\Omega}_{\Lambda_0}$ and $\overline{\Omega}_{\phi_0}$ be the closed linear span of $\Omega_{\Lambda_0}$ and  $\Omega_{\phi_0}$, respectively.
Define $\dot{\ell}_{\phi}(\btheta_0)[\bz]=(\dot{\ell}_{\phi}(\btheta_0)[z_1], \ldots, \dot{\ell}_{\phi}(\btheta_0)[z_p])^{\top}$ and $\dot{\ell}_{\Lambda}(\btheta_0)[\bq]=( \dot{\ell}_{\Lambda}(\btheta_0)[q_1], \ldots, \dot{\ell}_{\Lambda}(\btheta_0) [q_p])^{\top}$, where $\bz=(z_1, \ldots, z_p)^{\top}$  $ \in \overline{\Omega}_{\phi_0}^p$,  
$\bq=$ $(q_1, \ldots, q_p)^{\top} \in \overline{\Omega}_{\Lambda_0}^p$.

Referring to  Theorem 1 in \cite{bickel1993efficient}(pp. 70), under Conditions (C1)--(C3) and (C7), the efficient score vector for $\bbeta$ is defined as
$$
\ell_{\boldsymbol{\beta}}^*(\btheta_0)=\dot{\ell}_{\boldsymbol{\beta}}(\btheta_0)-\dot{\ell}_{\Lambda}(\btheta_0)[\bq^*]-\dot{\ell}_{\phi}(\btheta_0)[\bz^*],
$$
where $(\bq^{*\top}, \bz^{*\top})^{\top} \in \overline{\Omega}_{\Lambda_0}^p \times \overline{\Omega}_{\phi_0}^p$ represents the least favorable direction chosen
such that $	\ell_{\boldsymbol{\beta}}^*(\btheta_0)$ is orthogonal to  $\dot{\ell}_{\Lambda}(\btheta_0)[\bq^*]$ and $\dot{\ell}_{\phi}(\btheta_0)[\bz^*]$, respectively.
This involves solving the following normal equations to obtain $\bq^*$ and $\bz^*$:
\begin{equation*} \dot{\ell}_{\boldsymbol{\beta}}(\btheta_0)\dot{\ell}^*_{\Lambda}(\btheta_0)=(\dot{\ell}_{\Lambda}(\btheta_0)[\bq^*]+\dot{\ell}_{\phi}(\btheta_0)[\bz^*])\dot{\ell}^*_{\Lambda}(\btheta_0),
\label{lLam}
\end{equation*}
and
\begin{equation*} \dot{\ell}_{\boldsymbol{\beta}}(\btheta_0)\dot{\ell}^*_{\phi}(\btheta_0)=(\dot{\ell}_{\Lambda}(\btheta_0)[\bq^*]+\dot{\ell}_{\phi}(\btheta_0)[\bz^*])\dot{\ell}^*_{\phi}(\btheta_0),
\label{lphi}
\end{equation*}
where $ \dot{\ell}^*_{\Lambda}(\btheta_0)$ and $\dot{\ell}^*_{\phi}(\btheta_0)$ denote the adjoint operator of $\dot{\ell}_{\Lambda}(\btheta_0)$ and $\dot{\ell}_{\phi}(\btheta_0) $, respectively.
To prove the asymptotic normality of $ \hat{\bbeta} $,  we need to verify the following three conditions:
\begin{enumerate}[(1)] 
\item 
$ \bz^*$ and $\bq^*$ exist, which ensures  the existence of the efficient score $\ell_{\boldsymbol{\beta}}^*(\hat{\btheta})$;

\item 
the efficient score $\ell_{\boldsymbol{\beta}}^*(\hat{\btheta})$ belongs to a Donsker class  and converges to $	\ell_{\boldsymbol{\beta}}^*(\btheta_0)=\dot{\ell}_{\boldsymbol{\beta}}(\btheta_0)-\dot{\ell}_{\Lambda}(\btheta_0)[\bq^*]-\dot{\ell}_{\phi}(\btheta_0)[\bz^*]$ in the $L_2(\mathbb{P})$-norm;

\item 
the matrix $\mathbf{I}(\boldsymbol{\beta}_0) = E\{\ell_{\boldsymbol{\beta}}^*(\boldsymbol{\theta}_0 )\ell_{\boldsymbol{\beta}}^*(\boldsymbol{\theta}_0)^{\top}\}$ is nonsingular.
\end{enumerate}

For Condition (1), the existence of $ \bz^*$ and $\bq^*$ can be established through same arguments as presented in Theorem 6.1 of \cite{huang1997sieve}.
Next, we verify Condition (2). 
Similar to the proof of Theorem 2, for every $\delta>0$, we can conclude that the class $\{\dot{\ell}_{\bbeta}(\btheta_n): \btheta_n \in  \mathcal{D} \times  \mathcal{M}_{B} \times  \mathcal{F}_{B}, d(\btheta_n, \btheta_{0}) \leq \delta\} $ has bounded $\varepsilon$-bracketing number $M (\delta/\varepsilon)^{p}$ and is thus Donsker.
Similar arguments show that $\{\dot{\ell}_{\Lambda}(\btheta_n)[\bq^*]: \btheta_n \in  \mathcal{D} \times  \mathcal{M}_{B} \times  \mathcal{F}_{B}, d(\btheta_n, \btheta_{0}) \leq \delta\} $ and $\{\dot{\ell}_{\phi}(\btheta_n)[\bz^*]: \btheta_n \in  \mathcal{D} \times  \mathcal{M}_{B} \times  \mathcal{F}_{B}, d(\btheta_n, \btheta_{0}) \leq \delta\} $ are both Donsker class.
It then follows
from the preservation of the Donsker property that $\ell_{\bbeta}^*(\hat{\btheta})$ belongs
to a Donsker class. 
Based on Theorem 1, we can conclude that $\ell_{\bbeta}^*(\hat{\btheta})$ converges to $\ell_{\bbeta}^*(\btheta_0)$ in   $L_2(\mathbb{P})$-norm, implying that $\mathbb{G}_n \{\ell_{\boldsymbol{\beta}}^*(\hat{\btheta}) \}$ converges in distribution to a zero-mean $p$-variate normal random vector.
By the definition of $\hat{\btheta}$, we have $\mathbb{P}_n\{\ell_{\bbeta}^*(\hat{\btheta})\}=0$, it follows that
\begin{equation*}
\mathbb{G}_n\{\ell_{\bbeta}^*(\hat{\btheta})\}=- n^{1/2} \mathbb{P}\{\ell_{\bbeta}^*(\hat{\btheta})- \ell_{\bbeta}^*(\btheta_0)\}.
\label{Gn}
\end{equation*}  
By the Taylor series expansion, we have  
\begin{equation}\tag{S7} 
\begin{aligned} 	\mathbb{G}_n\{\ell_{\bbeta}^*(\hat{\btheta})\}=  n^{1/2}  \mathbf{I}(\boldsymbol{\beta}_0) (\hat{\boldsymbol{\beta}}-\boldsymbol{\beta}_0) + O(n^{1/2}  d^2(\hat{\btheta}, \btheta_0)).
\end{aligned}
\label{Taylor}
\end{equation}
By the convergence rate   $d(\hat{\boldsymbol{\theta}},\boldsymbol{\theta}_0) = O_p(\alpha_n\log^2n+n^{-\xi\nu})$ and assuming $(2\xi+1)^{-1}<\nu<(2\xi)^{-1}$ with $ \xi \geq 1$ and $ \sqrt{n}\alpha_n^2\to0$ as $n\to\infty$, it follows that $d^2(\hat{\boldsymbol{\theta}},\boldsymbol{\theta}_0)=o_p(n^{-1/2})$.

Finally, we show that $\mathbf{I}(\boldsymbol{\beta}_0) $ is nonsingular.
Suppose that the matrix $\mathbf{I}(\boldsymbol{\beta}_0)$ is singular, there exists a nonzero vector $\tilde{\veta}$ such that 
$$\tilde{\veta}^{\top} \mathbf{I}(\boldsymbol{\beta}_0)\tilde{\veta} = \tilde{\veta}^{\top} E\{\ell_{\boldsymbol{\beta}}^*(\boldsymbol{\theta}_0 )\ell_{\boldsymbol{\beta}}^*(\boldsymbol{\theta}_0)^{\top}\}\tilde{\veta}=0 .$$
This indicates that the score function along the submodel $\{\bbeta_{0}+\kappa \tilde{\veta},\Lambda_{0}+\kappa \tilde{\veta}^{\top} \bq^*, \phi_{0}+\kappa \tilde{\veta}^{\top} \bz^*\}$ is zero with probability 1, that is 
$$
\begin{aligned}
&Q_1(\bX,\bU,\bV ; \btheta_0)[\tilde{\veta}^{\top} \bX+ \tilde{\veta}^{\top} \bz^*(w) +\tilde{\veta}^{\top} \bq^*(R)/ \Lambda_0(R)]\\ 
&+Q_2(\bX,\bU,\bV ; \btheta_0)[\tilde{\veta}^{\top} \bX+ \tilde{\veta}^{\top} \bz^*(w) +\tilde{\veta}^{\top} \bq^*(L)/ \Lambda_0(L)] =0.
\end{aligned}
$$
We consider $  \delta_{L} =1 $, and 
then   have  $ Q_1(\bX,\bU,\bV ; \btheta_0)[\tilde{\veta}^{\top} \bX+ \tilde{\veta}^{\top} \bz^*(w) +\tilde{\veta}^{\top} \bq^*(R)/ \Lambda_0(R)]=0 $
Therefore, with probability 1,  $ \tilde{\veta}^{\top} \bX+ \tilde{\veta}^{\top} \bq^*(R)/\Lambda_0(R) + \tilde{\veta}^{\top} \bz^*(w)=0 $.
It then follows that $\tilde{\veta} =0 $ by Condition (C8).
This contradicts the fact that $\tilde{\veta}$ is a nonzero vector. 
Therefore,   $\mathbf{I}(\boldsymbol{\beta}_0)$ is nonsingular. 
By  (\ref{Taylor}), we have
$$	\sqrt{n}(\hat{\boldsymbol{\beta}}-\boldsymbol{\beta}_0)\stackrel{d}{\to}N(\boldsymbol{0},\boldsymbol{I}^{-1}(\boldsymbol{\beta}_0)), n\to\infty, $$
which completes  the proof of Theorem 3.
\\

\section*{Section S.3: Additional simulation results}

In this section, we conducted simulations under   the PO model.
In particular, the failure times of interest were generated from
model (1) with  $G(x) = \log(1 + x)$.
Other simulation configurations remained the same as the simulation study of the main manuscript.
The  results summarized in
Tables S1 and S2    suggest similar conclusions as Section 4 of the main manuscript 
regarding the comparisons of the proposed method,
``Spline-Trans''  and ``NPMLE-Trans''.
In addition, it is worth pointing out that the performances of ``deep PH" and ``penalized PH" deteriorate remarkably here due to using a misspecified PH model.

\begin{table}
\caption*{Table S1: Simulation results for the regression parameter estimates under the PO model. Results include the estimation bias (Bias),  the sample standard error (SSE) of the estimates,  the average of the standard error estimates (SEE),  and the 95\% coverage probability (CP95).}\label{table3}

\begin{center}
	\scalebox{0.68}{
	\begin{threeparttable}
	\begin{tabular}{rrrrrrrrrrrrrrrrrrr}
	&        &       & \multicolumn{4}{c} {Proposed method}   & & \multicolumn{2}{c}{Deep PH}      & & \multicolumn{2}{c}{Spline-Trans} & & \multicolumn{2}{c}{ Penalized PH}& & \multicolumn{2}{c}{NPMLE-Trans }\\
	\cline{4-7} \cline{9-10}\cline{12-13}\cline{15-16}\cline{18-19}
	Case   & $n$ &  par. & Bias&SSE & SEE&CP95& & Bias&SSE & & Bias&SSE & & Bias&SSE& & Bias&SSE \\
	\hline
	1        & 500 & $\beta_{1}$& 0.006&0.117 &0.126 &0.970 &&  -0.086   &  0.123   & &0.038& 0.136& &-0.138&0.097&&0.014& 0.120  \\
	&        &   $\beta_{2}$& -0.003&0.234  &0.239 &0.950  &&   0.109  &  0.252   & &-0.038& 0.253& &0.232&0.233&&-0.011& 0.237   \\
	&1000 &  $\beta_{1}$& 0.018&0.084 &0.087 &0.955 &&  -0.093   & 0.104    & & 0.035&0.090& &-0.110&0.066&&0.030& 0.086 \\
	&   &  $\beta_{2}$& -0.019&0.171 &0.166 &0.945 &&  0.094   & 0.184    & &-0.037& 0.181& &0.147&0.167&&-0.033& 0.174  \\
	2        & 500 &$\beta_{1}$& -0.022&0.111 &0.126 &0.985&&   -0.113  & 0.133      & &0.037& 0.127& &-0.171&0.098&&-0.036& 0.110 \\
	&     &   $\beta_{2}$& 0.015&0.232 &0.238 &0.975 &&   0.111  &    0.248 & &-0.037& 0.255& &0.262&0.235&&0.030& 0.237  \\
	&1000 &  $\beta_{1}$& -0.004&0.089 &0.087 &0.960 &&   -0.109  &  0.109    & &0.034& 0.093& &-0.145&0.070&&-0.018& 0.090\\
	&     & $\beta_{2}$& 0.005&0.170 &0.164 &0.945 &&   0.077  &   0.198  & &-0.036& 0.179& &0.190&0.169&&0.021& 0.170  \\  			
	3       & 500 &  $\beta_{1}$& 0.003&0.119 &0.125 &0.980&&   -0.081  &   0.138   & &0.033& 0.133& &-0.141&0.095&&-0.003& 0.118   \\
	&      &  $\beta_{2}$& 0.008&0.238 &0.237 &0.965 &&  0.082   &  0.286   & &-0.032& 0.265& &0.234&0.233&&0.012& 0.241  \\
	&1000 &$\beta_{1}$& 0.010&0.088 &0.086 &0.955 &&  -0.087   &   0.107  & &0.020& 0.092& &-0.116&0.069&&0.009& 0.089 \\
	&    &   $\beta_{2}$& -0.006&0.168 &0.163 &0.940 &&  0.074   & 0.185    & &-0.022& 0.178& &0.157&0.162&&-0.009& 0.170 \\
	4        & 500 &$\beta_{1}$& 0.033&0.128 &0.120 &0.924 &&   -0.113  &  0.121   & &0.114& 0.154& &-0.147&0.104&&0.023& 0.123\\
	&      &  $\beta_{2}$& -0.009&0.207 &0.222 &0.965&&  0.060   &  0.249   & &-0.086& 0.258& &0.248&0.201&&0.001& 0.204  \\
	&1000 &  $\beta_{1}$& $<$0.001&0.079 &0.081 &0.945 &&  -0.124   &  0.091   & &0.033& 0.087& &-0.157&0.058&&-0.006& 0.078  \\
	&     &  $\beta_{2}$& 0.002&0.159 &0.151 &0.955 &&   0.087  &   0.178  & &-0.031& 0.174& &0.194&0.139&&0.009& 0.155 \\
	5        & 500 & $\beta_{1}$& -0.001&0.128 &0.129 &0.945&&   -0.163  &  0.145   & &0.115& 0.167& &-0.184&0.109&&-0.024& 0.122 \\
	&   &  $\beta_{2}$&  0.015&0.231 &0.233 &0.955 &&  0.131   &  0.250   & &-0.099& 0.309& &0.281&0.223&&0.039& 0.227  \\
	&1000 &$\beta_{1}$& -0.022&0.089 &0.086 &0.940 &&  -0.194   &   0.100  & &0.038& 0.098& &-0.189&0.066&&-0.048& 0.083\\
	&    & $\beta_{2}$& 0.035&0.165 &0.155 &0.930&&   0.169  &  0.178   & &-0.029& 0.195& &0.249&0.157&&0.059& 0.155 \\ 		
	6       & 500 &  $\beta_{1}$&-0.002 & 0.133 & 0.125 & 0.919&&  -0.168   &   0.114  & &0.067& 0.161& &-0.198&0.103&&-0.048 & 0.123 \\
	&   &   $\beta_{2}$&  0.016 & 0.224 & 0.225 & 0.929 &&   0.124 &  0.246   & &-0.050& 0.269& &0.276&0.197&&0.057& 0.214  \\
	&1000 &  $\beta_{1}$&-0.029 & 0.079 & 0.083 & 0.954 &&   -0.164  &   0.094 & &-0.019& 0.085& &-0.210&0.054&&-0.082& 0.076 \\
	&    &   $\beta_{2}$& 0.039 & 0.154 & 0.147 & 0.929&&  0.096  &   0.197  & &0.010& 0.173& &0.251&0.144&&0.086& 0.147 \\ 	
					
	\end{tabular}
	\begin{tablenotes}
	\footnotesize
	\item[*]Note:  ``Proposed method'' denotes the proposed deep regression approach,   ``Deep PH '' refers to the deep learning approach  for the partially linear PH model \citep{du2024deep},  ``Spline-Trans'' refers to the sieve  MLE approach for the partially linear additive transformation model   \citep{Yuan2024SIM}, ``Penalized PH'' refers to penalized MLE method for the PH model with linear covariate effects  \citep{zhao2020simultaneous},     ``NPMLE-Trans'' refers to the nonparametric MLE method for the  transformation model   with linear   covariate effects   \citep{zeng2016maximum}. \\
			\end{tablenotes}
		\end{threeparttable}
		}
	\end{center}
\end{table}

\begin{table}
\caption*{Table S2: Simulation results for
	the relative error (RE) of  $\hat{\phi}_{\balpha}$ over  $\phi$ and the mean squared  error (MSE) of the estimated survival function.
	Values of  MSE in the table are multiplied by  100.}\label{table4}

	\begin{center}
	\scalebox{0.8}{
	\begin{threeparttable}
	\begin{tabular}{rrrrrrrrrrrrrrrr}
	&        &        \multicolumn{2}{c} {Proposed method}    & & \multicolumn{2}{c}{Deep PH}    & & \multicolumn{2}{c}{Spline-Trans} & & \multicolumn{2}{c}{ Penalized PH}& & \multicolumn{2}{c}{NPMLE-Trans }\\
	\cline{3-4} \cline{6-7}\cline{9-10}\cline{12-13}\cline{15-16}
	Case   & $n$ &   RE&MSE && RE&MSE&& RE&MSE& & RE&MSE & & RE&MSE \\
	\hline
	1        & 500  & 0.399&0.268  &&  0.565  &   0.829  & &2.285& 0.819& &0.435&0.378&&0.333& 0.549 \\
	&1000 & 0.284&0.131 &&  0.457  &   0.738  & &1.946& 0.393& &0.350&0.261&&0.240&  0.253 \\
	2        & 500 & 0.648&0.719 &&  0.792  &    1.416  & &1.407& 0.850& &0.887&1.169&&0.875& 1.469  \\
	&1000 & 0.519&0.448 &&  0.658  &  1.313   & &0.975& 0.404& &0.857&1.054&&0.849& 1.220  \\   			
	3       & 500 & 0.890&0.518  && 1.025   &  1.058   & &2.257&1.533& &0.996&0.656&&1.065& 0.983  \\
	&1000 & 0.719&0.315 &&  0.807  &  0.831   & &1.852& 1.026& &0.984&0.578&&1.014& 0.705  \\
	4        & 500 & 0.565&0.546 && 0.659  &  1.477   & &2.568& 1.831 & &0.519&0.651&&0.469& 0.791  \\
	&1000 & 0.401&0.281 &&   0.571   &  1.390  & &2.150& 0.793& &0.443&0.518&&0.325& 0.387  \\
	5        & 500 & 0.440&1.430 &&0.661  & 2.960& &1.631& 0.979& &0.535&1.555&&0.440& 1.949 \\
	&1000 & 0.368&0.954&&  0.617   &   2.720     & &1.459& 0.395& &0.519&1.380&&0.415& 1.420  \\
	6       & 500 & 0.653  & 1.638 && 0.837   &   2.834   & &1.526& 3.597& &0.807&2.029&&0.775& 2.583  \\
	&1000 & 0.568 & 1.130 && 0.752   &   2.557  & &1.237& 2.545& &0.798&1.864&&0.758& 2.117  \\  	
\end{tabular}
\begin{tablenotes}
\footnotesize
\item[*]Note:  ``Proposed method'' denotes the proposed deep regression approach,   ``Deep PH '' refers to the deep learning approach  for the partially linear PH model \citep{du2024deep},  ``Spline-Trans'' refers to the sieve  MLE approach for the partially linear additive transformation model   \citep{Yuan2024SIM}, ``Penalized PH'' refers to penalized MLE method for the PH model with linear covariate effects  \citep{zhao2020simultaneous},     ``NPMLE-Trans'' refers to the nonparametric MLE method for the  transformation model   with linear   covariate effects   \citep{zeng2016maximum}. \\
\end{tablenotes}
\end{threeparttable}
		}
	\end{center}
\end{table}

\section*{Section S.4:  More details about ADNI data and prediction evaluation}

In the ADNI data analysis, 
we focused on 14 potential risk factors for AD, consisting of a genetic covariate:
\texttt{APOE$\epsilon$4} (coded as 0, 1, or 2 based on the number of APOE 4 alleles),
four demographic covariates:
\texttt{Age} (given in years),
\texttt{Gender} (1 for male and 0 for female),
\texttt{Education} (years of education),
\texttt{Marry} (for married and 0 otherwise),
and nine test scores characterizing cognitive ability, memory and executive function:
\texttt{ADAS11} (a test score  measured by the AD Assessment Scale Cognitive Subscales (ADAS) 11),
\texttt{ADAS13} (a test score  measured by the ADAS13),
\texttt{MMSE} (a test score  measured by the Mini-Mental State Examination),
\texttt{RAVLT.i} (the immediate recall  in
the Rey Auditory Verbal Learning Test (RAVLT)),
\texttt{RAVLT.l} (learning ability in
the RAVLT),
\texttt{RAVLT.d} (30-minute delayed recall in
the RAVLT),
\texttt{DelayRec} (30-minute delayed recognition in
the RAVLT),
\texttt{TrailA} (a score obtained from the Trail Making Test Part A),
and \texttt{TrailB} (a score  obtained from the Trail Making Test Part B).
In particular,  \texttt{ADAS11},  \texttt{ADAS13}, and  \texttt{MMSE} collectively assess an individual's cognitive ability.
An individual with higher  ADAS11 and ADAS13 has a worse cognitive function,
whereas higher   MMSE represents a  better cognitive condition.
Memory performance is mainly measured through four scores from the RAVLT, with higher values of  \texttt{RAVLT.i},  \texttt{RAVLT.l},  \texttt{RAVLT.d}, and  \texttt{DelayRec} indicating better performance in immediate recall, learning ability, 30-minute delayed recall, and 30-minute delayed recognition, respectively.
Higher values for  \texttt{TrailA} and  \texttt{TrailB} reflect longer processing times to complete tasks in Test Part A and Part B,
respectively,  suggesting poorer executive function.
The continuous variables were all normalized to have a mean of 0 and a variance of 1.


To measure the prediction performance of the proposed method, we utilized an integrated Brier score (IBS) \citep{2015MeasuresOD} calculated on the test data.
Specifically, IBS is defined as
$$
{\rm IBS}(\hat{S}) = \frac{1}{n} \sum_{i=1}^{n}\frac{1}{\widetilde{\tau}} \int_{0}^{\widetilde{\tau}} \{I(T_i>t \mid \bX_i,\bW_i)- \hat{S}(t\mid \bX_i,\bW_i)\}^2 dt,
$$
where $\widetilde{\tau}$ represents  the maximum finite value of all observed $ \{L_i,R_i; i = 1, \ldots, n\}$,
and  $\hat{S}(t \mid \bX_i, \bW_i) =  \exp ( -  G[\hat{\Lambda} (t)    \exp  \{  \hat{\bbeta} ^{\top}  \bX_i + \hat{\phi}(\bW_i)\}] )$. Since $T_i \in [L_i, R_i) $ for each  $i = 1, \ldots, n$,
we clearly    have $I(T_i>t \mid \bX_i,\bW_i)=0$ when $t > R_i$ and
$I(T_i>t \mid \bX_i,\bW_i)=1$ if $t \leq L_i$.
When $L_i< t \leq R_i$, the exact value of $I(T_i>t \mid \bX_i,\bW_i)$ is unknown, and we can approximate it with $\hat{I}(T_i>t \mid \bX_i,\bW_i)=\{\hat{S}(t\mid \bX_i,\bW_i)-\hat{S}(R_i\mid \bX_i,\bW_i)\}/\{\hat{S}(L_i\mid \bX_i,\bW_i)-\hat{S}(R_i\mid \bX_i,\bW_i)\}$.
For the special case of $L_i<t \leq R_i = \infty$, $\hat{I}(T_i>t \mid \bX_i,\bW_i)$ equals $\hat{S}(t\mid \bX_i,\bW_i)/\hat{S}(L_i\mid \bX_i,\bW_i)$. A smaller value of IBS corresponds to a better prediction performance.

To investigate the prediction performance of the proposed deep method,  we conducted a 5-fold cross-validation, where the IBS was calculated with  20\% of the data in each fold, and the remaining  80\% of the data were treated as the training data used to estimate the  model parameters.
Figure S1 presents the obtained IBS values across all folds under the proposed deep regression method with the optimal model ($r=2.6$)
and the  comparative  methods.
It shows that the proposed method leads to the smallest IBS value except for the first fold,
confirming our method's powerful prediction ability.

\begin{figure}
	\begin{center}
		\includegraphics[scale=0.7]{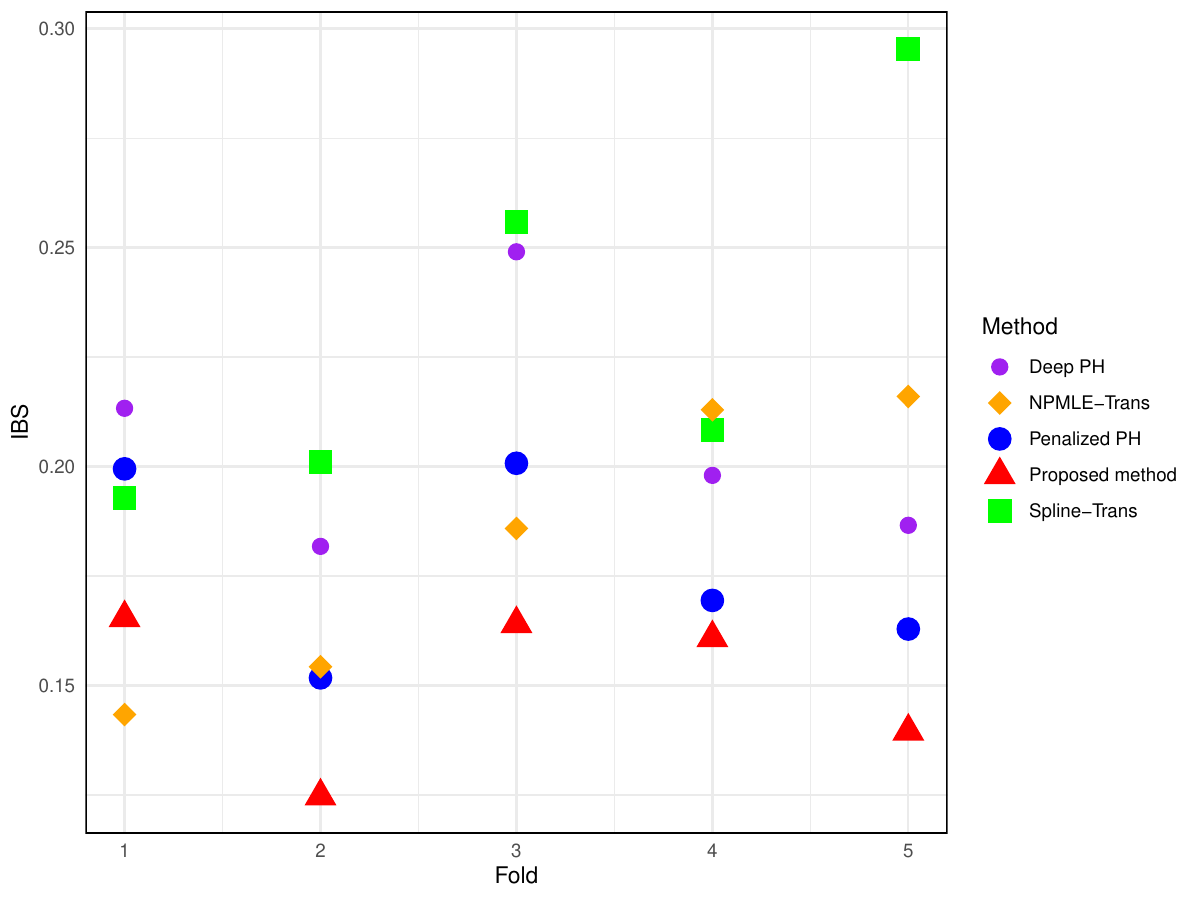}
	\end{center}
	\caption*{Figure S1: IBS values obtained from the ADNI data analysis. \label{fig3}}
\end{figure}

%